\documentclass{article}
\pdfoutput=1

\usepackage[final]{corl_2019} 
\usepackage[ruled,vlined]{algorithm2e}
\usepackage{algpseudocode}
\usepackage{amsfonts}
\usepackage{amsmath}

\makeatletter
\newcommand{\algorithmfootnote}[2][\footnotesize]{%
  \let\old@algocf@finish\@algocf@finish
  \def\@algocf@finish{\old@algocf@finish
    \leavevmode\rlap{\begin{minipage}{\linewidth}
    #1#2
    \end{minipage}}%
  }%
}
\makeatother

\usepackage{graphicx}
\graphicspath{ {./images/} }

\title{Dynamic Experience Replay}

\author{
  Jieliang Luo and Hui Li\\
  Autodesk Research, San Francisco, United States\\
  \texttt{rodger.luo@autodesk.com}, \texttt{hui.xylo.li@autodesk.com} \\
}

\begin{document}
\maketitle


\begin{abstract}
We present a novel technique called Dynamic Experience Replay (DER) that allows Reinforcement Learning (RL) algorithms to use experience replay samples not only from human demonstrations but also successful transitions generated by RL agents during training and therefore improve training efficiency. It can be combined with an arbitrary off-policy RL algorithm, such as DDPG~\citep{lillicrap2016ddpg} or DQN~\citep{mnih2015dqn}, and their distributed versions.\\ 


We build upon Ape-X DDPG~\citep{horgan2018distributed} and demonstrate our approach on robotic tight-fitting joint assembly tasks, based on force/torque and Cartesian pose observations. In particular, we run experiments on two different tasks: peg-in-hole and lap-joint. In each case, we compare different replay buffer structures and how DER affects them. Our ablation studies show that Dynamic Experience Replay is a crucial ingredient that either largely shortens the training time in these challenging environments or solves the tasks that the vanilla Ape-X DDPG cannot solve. We also show that our policies learned purely in simulation can be deployed successfully on the real robot. The video presenting our experiments is available at \url{https://sites.google.com/site/dynamicexperiencereplay} 
\end{abstract}



\section{Introduction}

Industrial robots have been heavily used in manufacturing and other industries, however, as they rely on pre-defined trajectories, they require precise calibration and fail to adapt to uncertainties. Adaptability to imprecision, varying conditions, and less structured environments is key to the future of automation. Reinforcement Learning (RL) has recently led to a range of successes in solving sequential decision-making problems, including learning control policies for robotic tasks. The control policies are learned through agents interacting with their surrounding environments and hold promises for generalizing to new scenarios in reaction to real-time observations~\citep{sutton2018reinforcement, kober2013survey}.

We focus on robotic assembly tasks that involve contact forces, because such tasks are widespread in industrial applications and yet challenging for robots to do. When the assembly pieces are in contact with one another, pose observations (direct from motion capture or indirect from perception learning models) alone are often insufficient. We explicitly consider force/torque observations for policy learning. During training, we randomize the initial condition within a pre-defined range and show flexibility of the learned policy to varying conditions. 

Most of the recent success in RL was achieved using model-free methods~\citep{levine2013guided,mnih2015dqn,lillicrap2016ddpg,schulman2015trust,schulman2017proximal}. They tend to achieve optimal performance, are generally applicable, and are easy to implement, but it is achieved at the cost of being data intensive. Leveraging human demonstrations~\citep{vevcerik2017leveraging} as well as various experience replay~\citep{lin1992self,schaul2015prioritized,andrychowicz2017hindsight} has shown to improve data efficiency.

We present a novel technique called Dynamic Experience Replay (DER) that allows RL algorithms to use experience replay samples not only from human demonstrations but also successful transitions generated by RL agents during training and therefore improve training efficiency. It can be combined with an arbitrary off-policy RL algorithm, such as DDPG or DQN, and their distributed versions. DER can be seen as a technique of over-sampling the under-represented class (successful trajectories in our case) from an imbalanced dataset, which has been studied and addressed in supervised learning~\citep{chawla2002smote, he2008adasyn}. 

We build upon Ape-X DDPG and demonstrate our approach on robotic tight-fitting joint assembly tasks, in particular, peg-in-hole and lap-joint tasks. In each case, we compare different replay buffer structures and how DER affects them. Our ablation studies show that Dynamic Experience Replay is a crucial ingredient that largely shortens the training time in these challenging environments or solves the tasks that the vanilla Ape-X DDPG cannot solve. We also show that our policies learned purely in simulation can be deployed successfully on an industrial robotic arm performing the physical tasks.
 
The remainder of this paper is structured as follows. The problem statement and related work are stated in Sec.~\ref{sec:relates}, followed by a detailed explanation of the proposed Dynamic Experience Replay in Sec.~\ref{sec:DER}. Experiment setup, results, and deployment on a real robot are presented in Sec.~\ref{sec:experiments}. Sec.~\ref{sec:conclusion} concludes the paper and proposes future work. 


\section{Problem Statement and Related Work}
\label{sec:relates}

\subsection{Problem Statement}

The RL problem at hand can be described as learn an optimal policy $\pi_\theta(a_t|s_t)$ for choosing an action $a_t$ given the current observation $s_t$ in order to minimize the expected total loss:

\begin{equation}
\label{eq1}
\min_{\pi_\theta}\mathbb{E}_{\tau\sim\pi_\theta}(l(\tau))
\end{equation}

where $\theta$ is the parameterization of policy $\pi$, trajectory $\tau=\{s_0, a_0, s_1, a_1, ..., s_T, a_T\}$, $\pi_\theta(\tau)=p(s_0)\prod_1^T p(s_t|s_{t-1},a_{t-1})\pi_\theta(a_t|s_t)$, and $l$ is the loss function of the trajectory $\tau$.

Equation~\ref{eq1} can be solved if a dynamics model $p(x_t|x_{t-1},a_{t-1})$ is provided, however, the dynamics model in contact-rich tasks is difficult to obtain. Alternatively, the equation can be solved by model-free RL algorithms to avoid using dynamics. DDPG is a model-free off-policy RL algorithm
for continuous action spaces. In DDPG, an actor policy $\pi: S\rightarrow A$ is created to explore the space and store the collected transition $(s_j, a_j, s_{j+1}, r_j)$ in a replay buffer $R$. Meanwhile, a critic policy $Q: S \times A \rightarrow \mathbb{R}$ is created to approximate the actor's action-value function $Q^\pi$. 

We would like to learn an optimal policy, which takes Cartesian pose and force/torque observations as input and outputs Cartesian velocity.

\subsection{RL for High Precision Assembly}

RL has been studied actively in the area of high precision assembly as it can reduce human involvement and increase the robustness to uncertainties. Inout el al.~\citep{inoue2017deep} used a Q-learning based method with LSTM~\citep{hochreiter1997long} for Q-function approximation to solve low-tolerance peg-in-hole tasks. Luo el at.~\citep{luo2018deep} extended a model-based approach MDGPS~\citep{montgomery2016guided} with haptic feedback for learning the insertion of a peg into a deformable hole. Fan el at.~\citep{fan2018guided} combined DDPG~\citep{lillicrap2016ddpg} and GPS~\citep{levine2013guided} to take advantage of both model-free and model-based RL~\citep{sutton2018reinforcement} to solve high-precision Lego insertion tasks. Luo el at.~\citep{luo2019reinforcement} combines iLQG~\citep{todorov2005generalized} with force/torque information by incorporating an operational space controller to solve a group of high-precision assembly tasks. In our work, we use torque/force and pose in Cartesian space as observations and 6 DOF Cartesian velocities as actions. This method bypasses the robot dynamics, which are usually inaccurate in simulation.      

\subsection{Leveraging Experience Replay in RL}

Experience replay~\citep{lin1992self} has been used to improve training efficiency in many RL algorithms, particularly for model-free RL, as it's less sample efficient than model-based RL. The technique has become popular after it was incorporated in the DQN~\citep{mnih2015dqn} agent playing Atari games. Prioritized experience replay~\citep{schaul2015prioritized} is a further improvement to prioritize transitions so agents can learn from the most "relevant" experiences. Hindsight experience replay~\citep{andrychowicz2017hindsight} stored every transition in the replay buffer not only with the original goal but also with a subset of other goals to acquire a more generalized policy. DDPG from Demonstrations~\citep{vevcerik2017leveraging} modified DDPG to permanently store a set of human demonstrations in the replay buffer to solve a group of insertion tasks. We extend DDPG from Demonstrations to a distributed framework and propose a set of experience replay structures in the context of distributed RL. Details are discussed in Sec.~\ref{subsec: buffer_strcture}.      

\subsection{Distributed RL}

Distributed RL can greatly increase training efficiency of model-free RL. Ape-X~\citep{horgan2018distributed} disconnects exploration from learning by having multiple actors interact with their own environments and select actions from a shared neural network. D4PG~\citep{barth2018distributed}, with the Ape-X framework, uses a distributional critic update to achieve a more stable learning signal. There are also a growing number of examples applying the distributed architecture to popular RL algorithms, such as Distributed PPO~\citep{heess2017emergence} and Distributed BA3C~\citep{adamski2018distributed}. Since our action space is continuous, we build our algorithm based on RLlib's~\citep{liang2017rllib} implementation of Ape-X DDPG.       

\section{Method}
\label{sec:DER}

\begin{figure}[t]
    \centering
    \includegraphics[width=0.8\textwidth]{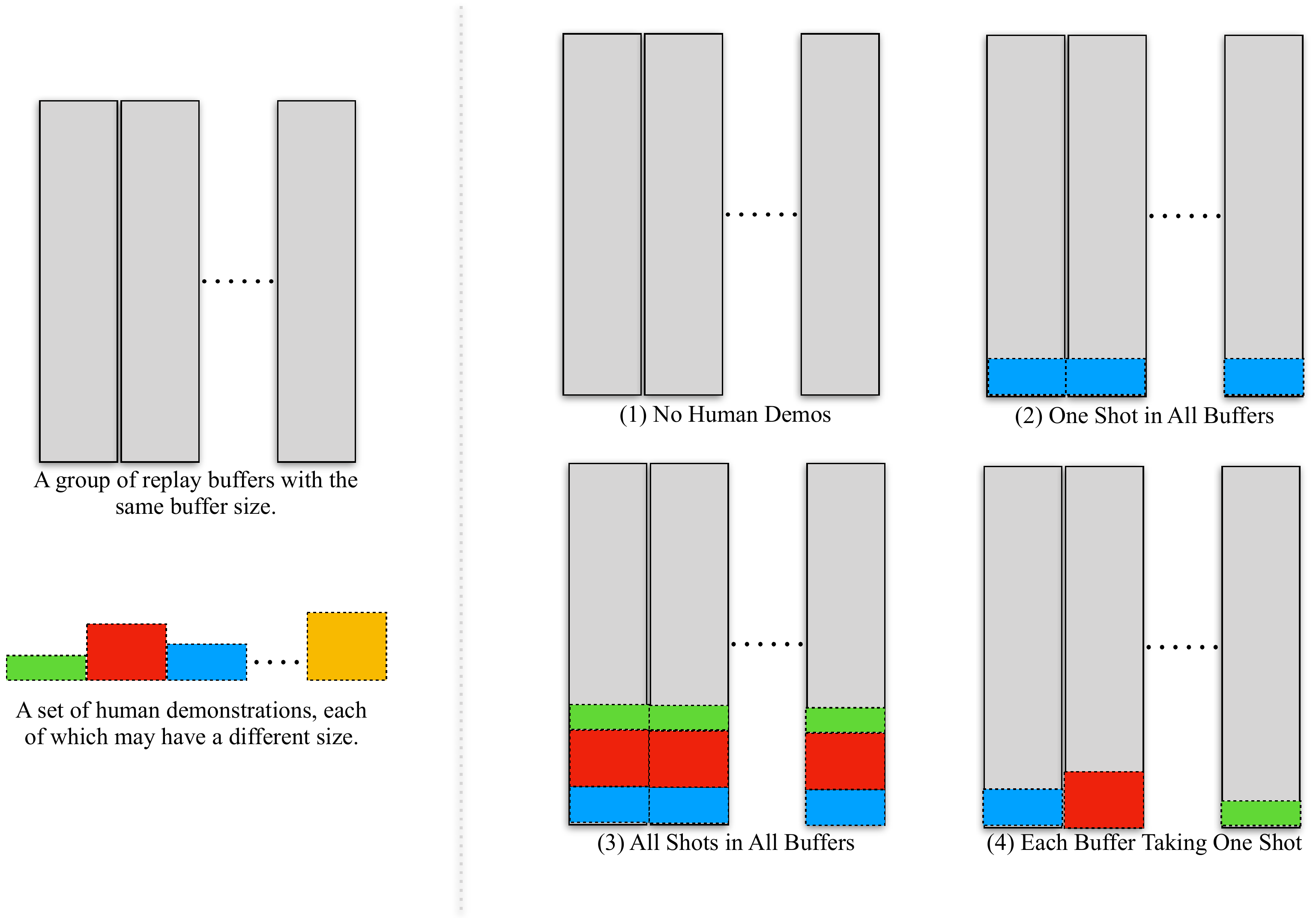}
    \caption{The four types of replay buffer structure for our experiments: (1) No human demonstrations in any buffer; (2) Same one-shot human demonstration in all buffers; (3) All human demonstrations in all buffers; (4) Each buffer with a different one-shot demonstration.}
    \label{fig:types}
\end{figure}

As model-free RL algorithms require excessive data, one or multiple shots of human demonstrations are sometimes introduced in the replay buffer $R$ for complex manipulation tasks~\citep{vevcerik2017leveraging, nair2018overcoming}. However, human demonstrations are not always helpful if the observation space during human demonstration does not match that during training. For example, for the high-precision lap-joint assembly task, we do not have haptic feedback during demonstration in simulation and only visual inspection is used, while during training, force/torque observations are required. Therefore, we propose a novel technique that augments human demonstrations with successful transitions generated by RL agents to improve training efficiency.    

\subsection{Setup}
\textbf{Observations:} The observation space is 13-dimensional. The policy is given as input the position ($x,y,z$) and orientation ($q_x,q_y,q_z,q_w$) of the timber piece attached to the robot end-effector, and the torque/force reading ($f_x,f_y,f_z,t_x,t_y,t_z$) from the sensor, which is mounted on the end of the robot arm. We do not use visual input to simplify the problem. 

\textbf{Actions:} The action space is 6-dimensional. The policy outputs the desired linear velocity ($v_x,v_y,v_z$) and angular velocity ($w_x,w_y,w_z$) of the timber piece attached to the robot end-effector.  

\textbf{Human demonstrations:}
For each task, depending on the replay buffer structure (Sec.~\ref{subsec: buffer_strcture}), zero, one or six human demonstrations are recorded in simulation, using a game controller to drive the robot end-effector until the joint is successfully assembled. Each demonstration includes all transitions from one successful episode. Each transition is of the form $e_t = (s_t, a_t, s_{t+1}, r_t)$.

\textbf{Rewards:} We use a simple linear reward function based on the distance between the goal pose and the current pose of the timber piece attached to the robot arm for both tasks. Additionally we use a large positive reward (+1000 for the peg-in-hole and +100 for the lap-joint) if the object is within a small distance of the goal pose: 
\[ 
r= \left \{
  \begin{tabular}{ccc}
  $-| g - x |$, & $| g - x | > \epsilon$ \\
  $-| g - x | + R$, & $| g - x | \leq \epsilon$ 
  \end{tabular}
  \right.
\]
where $x$ is the current pose of the object, $g$ is the goal pose, $\epsilon$ is a distance threshold, and $R$ is the large positive reward. We use negative distance as our reward function to discourage the behavior of loitering around the goal because the negative distance also contains time penalty.  

\subsection{Replay Structures in Distributed RL}
\label{subsec: buffer_strcture}

\begin{figure}[t]
    \centering
    \includegraphics[width=0.8\textwidth]{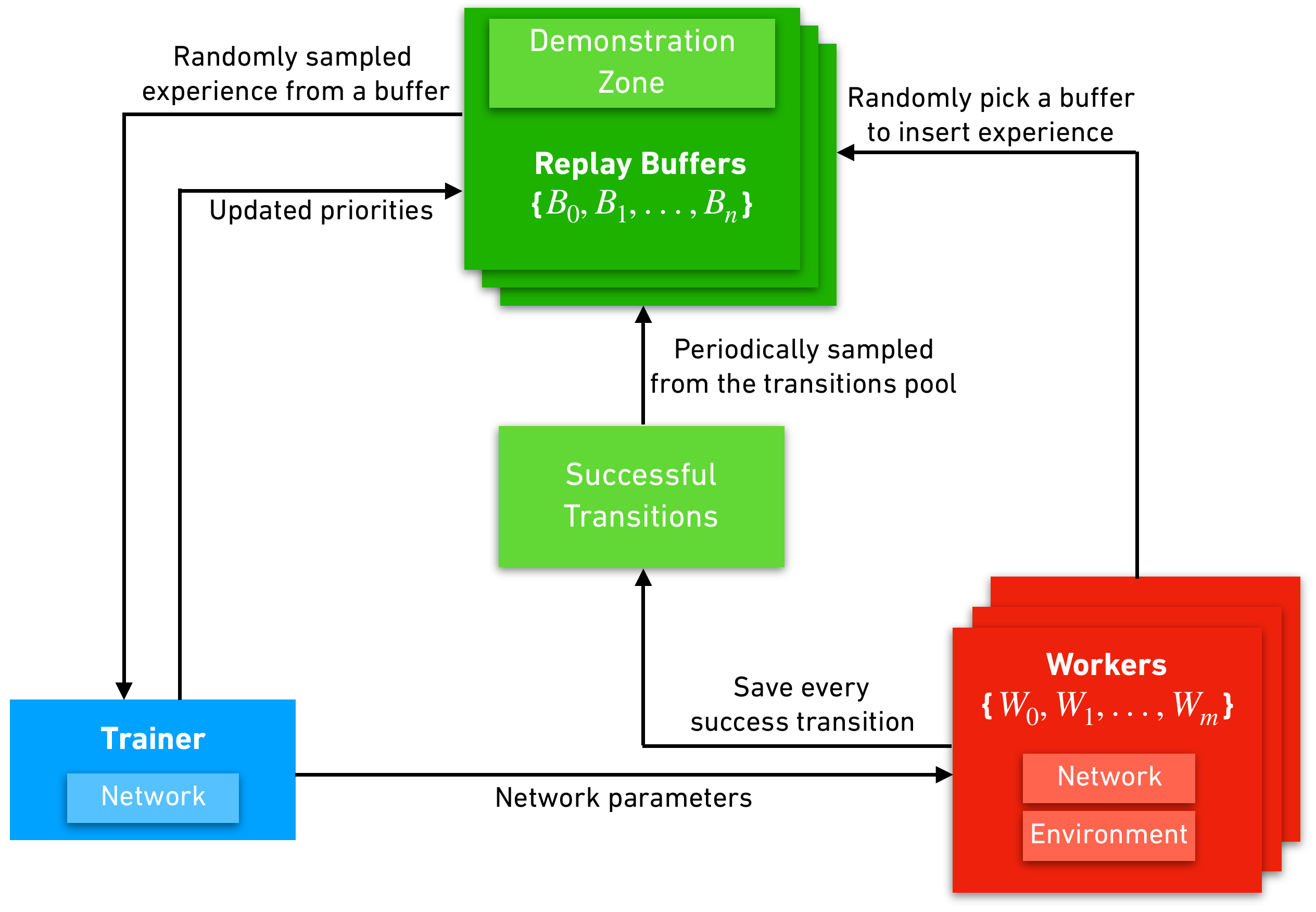}
    \caption{The Dynamic Experience Replay framework: multiple workers, each with its own instance of environment, and multiple replay buffers, each with capacity $\mathbb{C}$ for demonstrations. Human demonstration(s) are stored in the demonstration zones before training starts. During training, all successful transitions that are generated by workers are saved in a pool, which is sampled periodically by each replay buffer and stored in the demonstration zone.}
    \label{fig:structure}
\end{figure}

Off-policy RL algorithms perform experience replay by sampling minibatches from a pool of stored samples, which allows the use of arbitrary data like human demonstrations. Based on prioritized experience replay and Ape-X DDPG, we suggest four different replay buffer structures that can take advantage of demonstrations in distributed RL, as shown in Fig.~\ref{fig:types}.

Each buffer structure consists of a fixed number of replay buffers that load zero, one, or multiple human demonstrations before training starts. Without Dynamic Experience Replay, each buffer permanently keeps all the demonstrations with top priorities during training. The following section discusses how the buffer structures work with Dynamic Experience Replay. 

\subsection{Dynamic Experience Replay}
\label{subsec: dynamic}

The idea behind Dynamic Experience Replay (DER) is to augment human demonstrations using successful trajectories generated by RL agents during training, especially in cases where human demonstrations are not very helpful. We define \textit{demonstrations} as either human demonstrations or the successful trajectories generated by RL agents. If DER is activated, regardless of the buffer structures mentioned above, each buffer allocates capacity $\mathbb{C}$ specifically for demonstrations. We refer to this as the \textit{demonstration zone}. During training, all the successful episodes generated by RL agents are stored in a pool. Periodically, each replay buffer randomly samples one successful episode from the pool and stores it in the demonstration zone. When the demonstration zone is full, the oldest transitions are discarded. DER's framework in a distributed architecture is shown in Fig.~\ref{fig:structure}.

As in Ape-X, the DER algorithm consists of two concurrent parts, which are workers and a trainer. For each worker, after collecting the transitions of one episode, it randomly chooses a replay buffer and sends over the transitions. The trainer, in parallel, randomly selects a replay buffer and samples a batch of transitions for network update. The trainer also updates the priority of transitions in the selected buffer at the end of the training cycle. See Alg.~\ref{alg: der} for a formal description of the algorithm. 

\begin{algorithm}[h]
\label{alg: der}
\DontPrintSemicolon 

\textbf{Given:}\\

\begin{itemize}
  \item a distributed off-policy RL algorithm $\mathbb{A}$, \algorithmiccomment{$e.g.$ APE-X DDPG, APE-X DQN}
  \item an experience replay structure $\mathbb{S}$ $^1$.
  \item one-shot or a group of human demonstrations $\mathbb{D}$ (optional)
\end{itemize}

Initialize $\mathbb{A}$ \algorithmiccomment{Initialize neural networks}\\

Initialize replay buffers $\mathbb{B}$ \algorithmiccomment{Initialize a group of replay buffers}\\
Load $\mathbb{D}$ to $\mathbb{B}$ based on $\mathbb{S}$ \algorithmiccomment{Load human demos to the replay buffers based on the replay structure}\\
Initialize $\mathbb{T}$ \algorithmiccomment{Initialize a pool to save success transitions from agents}\\

\vspace{5mm}
\textbf{For each worker:}

\For{episode = $1$, $\mathbf{M}$ } {
    $\theta_0$ $\leftarrow$ Trainer.parameters() \algorithmiccomment{Update the latest network parameters for the trainer}\\
    s$_0$ $\leftarrow$ Environment.reset() \algorithmiccomment{Get initial state from its own environment}\\
    \For{ t = $1$, $\mathbf{T}$} {
        $a_{t-1}$ $\leftarrow$ $\pi_{\theta_{t-1}}$ $(s_{t-1})$ \algorithmiccomment{Choose an action from the current policy}\\
        ($r_{t-1}$, $s_t$) $\leftarrow$ Environment.step($a_{t-1}$) \algorithmiccomment{Apply the action to the environment}\\
        Transitions.Add($[s_{t}, a_{t-1}, r_{t-1}, s_{t-1}]$) \algorithmiccomment{Add data to a temp buffer}\\
    }
    $\mathbb{B}_n$.Add(Transitions) \algorithmiccomment{Send the transitions to a randomly selected replay buffer} \\
    \If{$episode_t$ succeeds}{ \algorithmiccomment{Save success transitions}\\ 
        $\mathbb{T}$.Add(Transitions) }
    Periodically($\theta_t$ $\leftarrow$ Trainer.Parameters()) \algorithmiccomment{Update to the latest network parameters}
   }
 
\vspace{5mm}
\textbf{For the trainer:}

$\theta_0$ $\leftarrow$ InitializeNetwork()\\
\For{o = $1$, $\mathbf{O}$}  
{   \algorithmiccomment{Update the parameters $O$ times}\\
    $\tau$ $\leftarrow$ $\mathbb{B}_n$.Sample() \algorithmiccomment {Sample a batch of transitions from a randomly selected buffer}\\
    $l_o$ $\leftarrow$ ComputeLoss($\tau; \theta_o$) \algorithmiccomment{Calculate loss using an off-policy algorithm, like DDPG}\\
    $\theta_{o+1}$ $\leftarrow$ UpdateParameters($l_o; \theta_o$)\\
    $p$ $\leftarrow$ ComputePriorities() \algorithmiccomment{Calculate priorities of the transitions in buffers$^2$}\\
    $\mathbb{B}_n$.SetPriority($p$) \algorithmiccomment{Update priorities to the selected buffer}\\
    Periodically($\mathbb{B}_i$.Update($\mathbb{T}_j$)) \algorithmiccomment{Replace previous demos with a success transition from the pool} 
}

\caption{Dynamic Experience Replay}

\algorithmfootnote{1. The four different experience replay structures are discussed in Sec~\ref{subsec: buffer_strcture}\\
2. We use absolute TD error for the calculation.}

\end{algorithm}

A hyper-parameter to experiment with DER is which replay buffer structure to use. In the next section, we compare the four types of replay buffer structure discussed in Sec.~\ref{subsec: buffer_strcture} and how DER affects them.

\section{Experiments}
\label{sec:experiments}	

This section is organized as follows. In Sec. \ref{sec: envs} we introduce distributed RL environments we use for the experiments as well as our training setup and procedure. In Sec. \ref{sec: result} we compare the performance of different replay buffer structures with and without DER. In Sec. \ref{sec: realRobot} we describe the deployment on the physical robot. 

The video presenting our experiments is available at \url{https://sites.google.com/site/dynamicexperiencereplay}. 

\subsection{Environments}
\label{sec: envs}

\begin{figure}[t]
    \centering
    \includegraphics[width=1.0\textwidth]{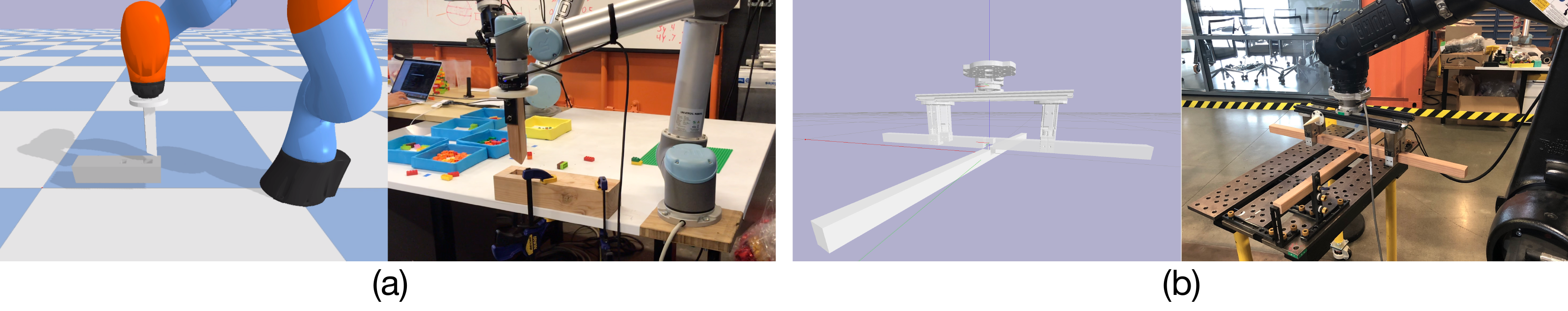}
    \caption{Two joint assembly tasks for algorithm evaluation: (a) chamfered peg-in-hole, (b) lap-joint. For both joints, the CAD model used in simulation is used to fabricate the real-world pieces.}
    \label{fig:tasks}
\end{figure}

We modeled our assembly tasks in the PyBullet~\citep{coumans2016pybullet} simulation engine. Specifically, we customized two tasks, chamfered peg-in-hole and lap-joint, which correspond to the real-world setup, as shown in Fig.~\ref{fig:tasks}.  

For the peg-in-hole task, we used a KUKA LRB iiwa robotic arm in simulation and attached a torque/force sensor between the end of the arm and a peg, as shown in Fig.~\ref{fig:tasks}(a). In order to be robot agnostic, we limit both the observations and actions in the Cartesian space. This way the trained model can be deployed on any arbitrary robotic arm. To demonstrate the point, we created a robot-less mode for training in simulation for the lap-joint task, as shown in Fig.~\ref{fig:tasks}(b). The robot-less setup helps us bypass needing a robot model in simulation, as most of them are inaccurate. 

For each task, we initialized 6 replay buffers and collected 6 human demonstrations. Each demonstration consists of a sequence of transitions of different lengths. Depending on which replay buffer structure is activated, the human demonstration data are used differently, as described in Fig.~\ref{fig:types}. Training is performed using the Ape-X DDPG algorithm and we adapted it from RLlib's implementation. 

\textbf{Initial states:} For the peg-in-hole tasks, the initial angle along the z-axis of the peg is randomized from 0 to 360 degree and other parts are fixed. For the lap-joint tasks, we randomized the initial angle along the z-axis and x-y position of the timber on the ground within a small range. The details of the initial randomnesses are documented in Fig.~\ref{fig:results_peg} and Fig.~\ref{fig:results_lap}.

\subsection{Results}
\label{sec: result}

\begin{figure}[h]
    
    \centering
    
    \begin{minipage}{0.38\textwidth}
        \centering
        \includegraphics[width=\textwidth]{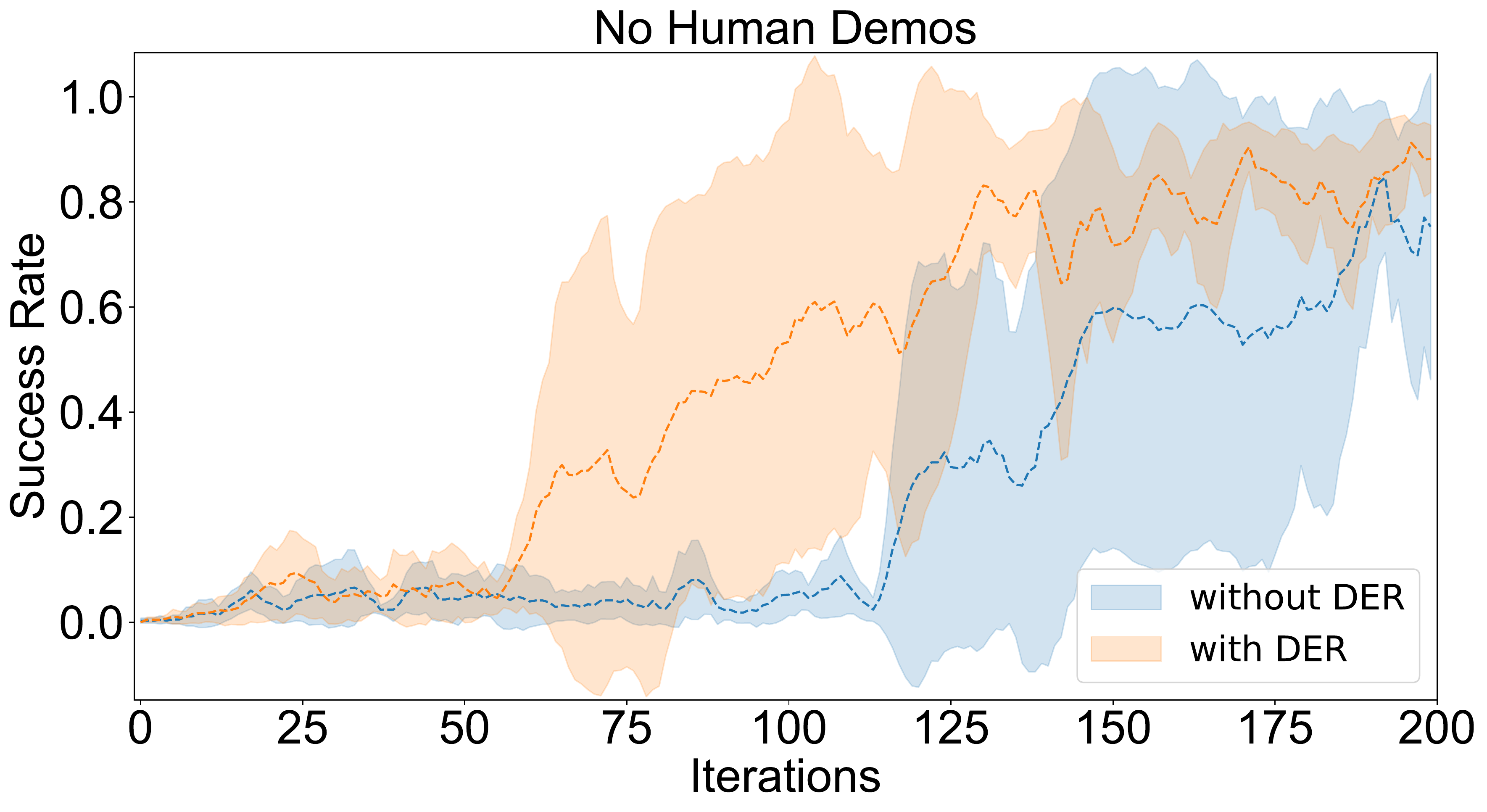} 
    \end{minipage}
    \begin{minipage}{0.38\textwidth}
        \centering
        \includegraphics[width=\textwidth]{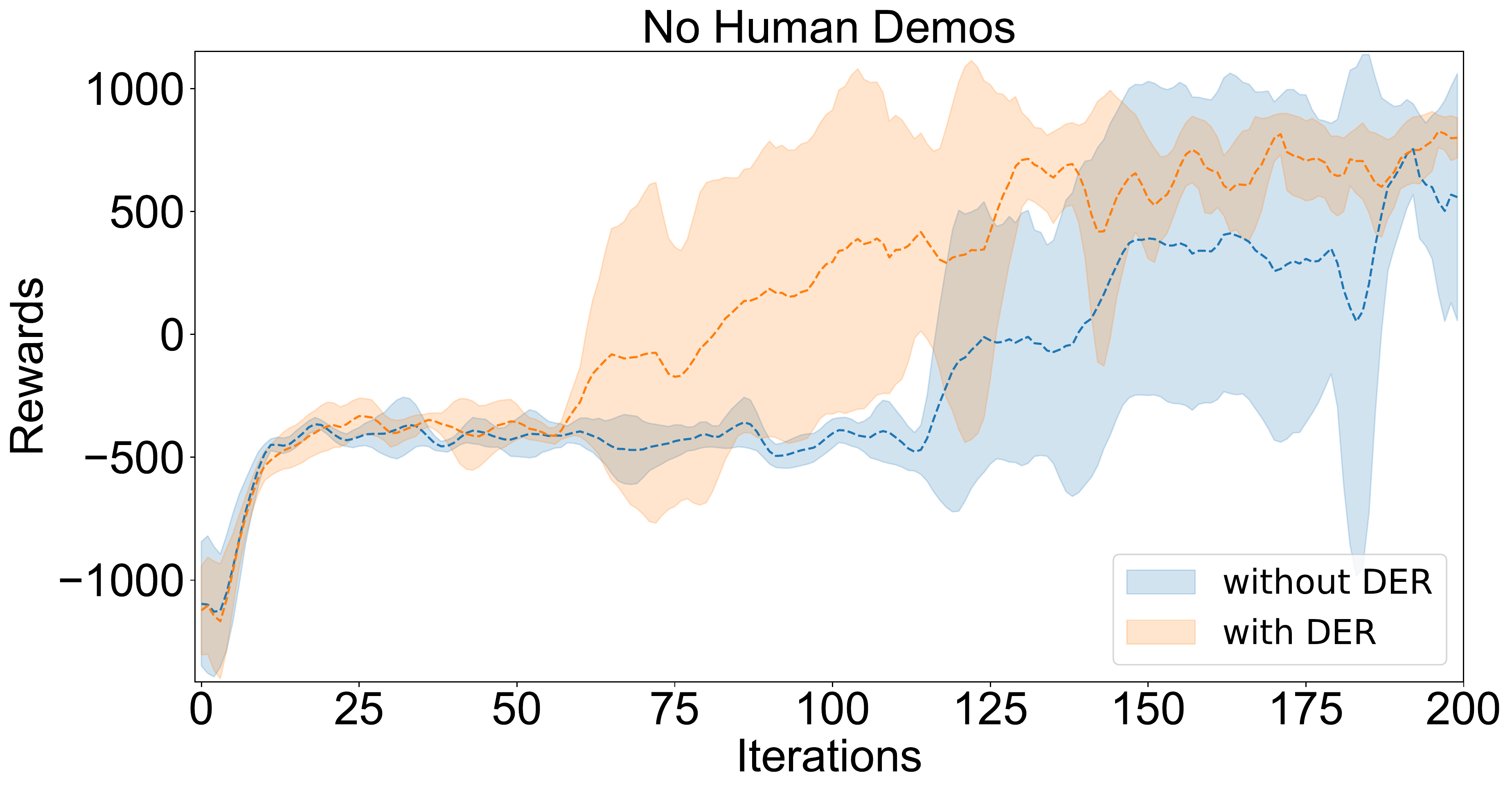} 
    \end{minipage}
    
    \begin{minipage}{0.38\textwidth}
        \centering
        \includegraphics[width=\textwidth]{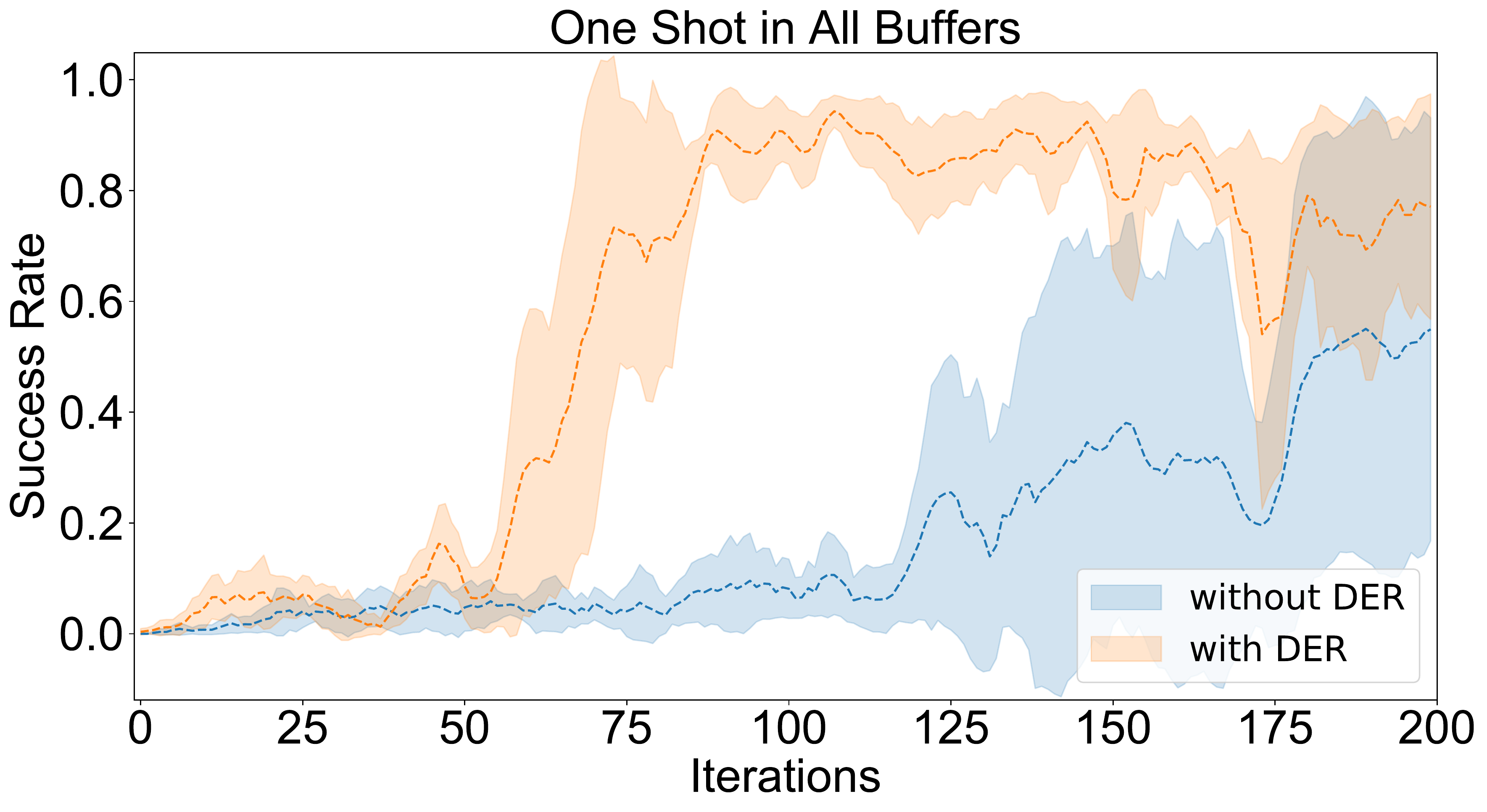} 
    \end{minipage}
    \begin{minipage}{0.38\textwidth}
        \centering
        \includegraphics[width=\textwidth]{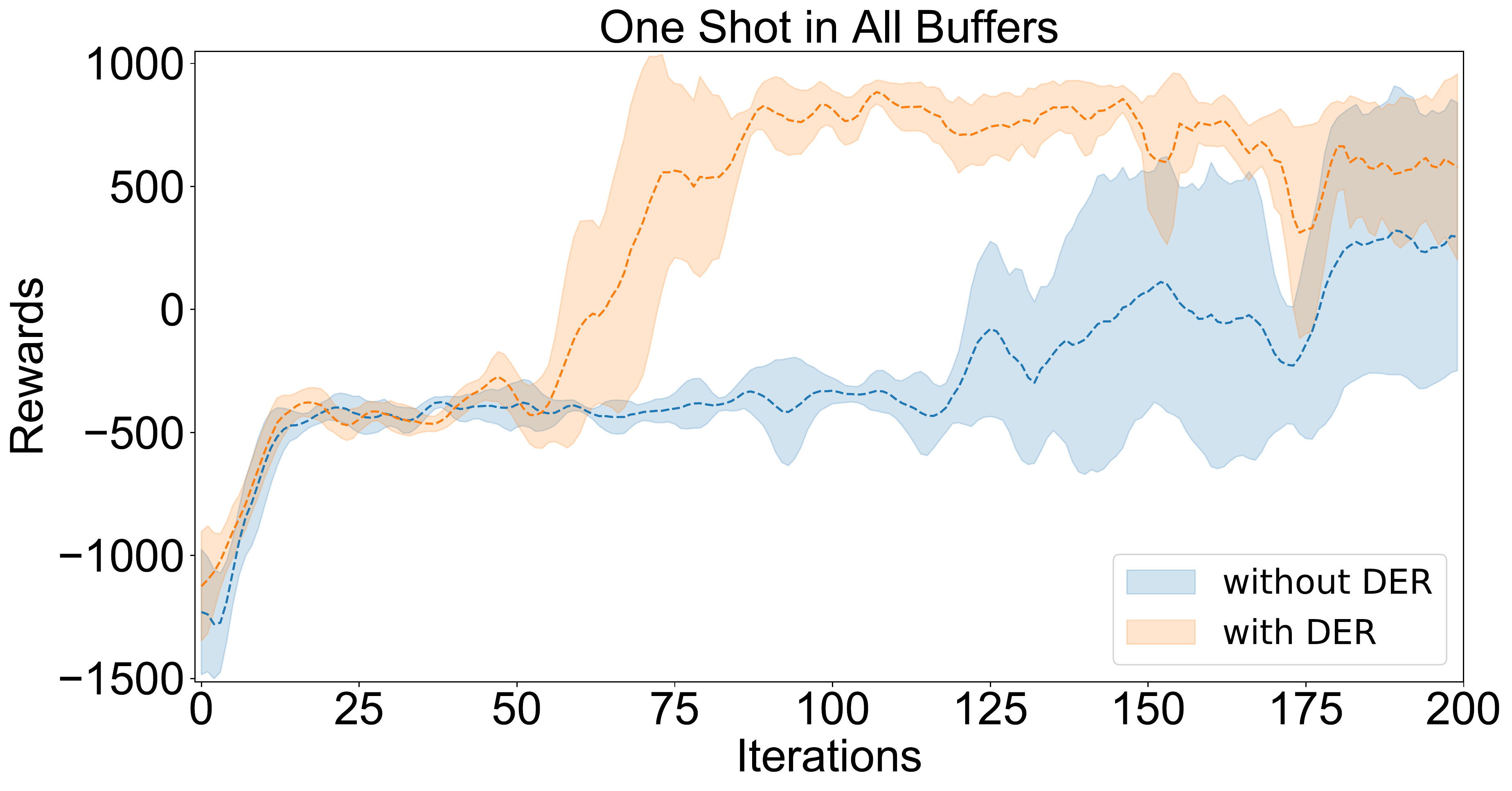}
    \end{minipage}
    
    \begin{minipage}{0.38\textwidth}
        \centering
        \includegraphics[width=\textwidth]{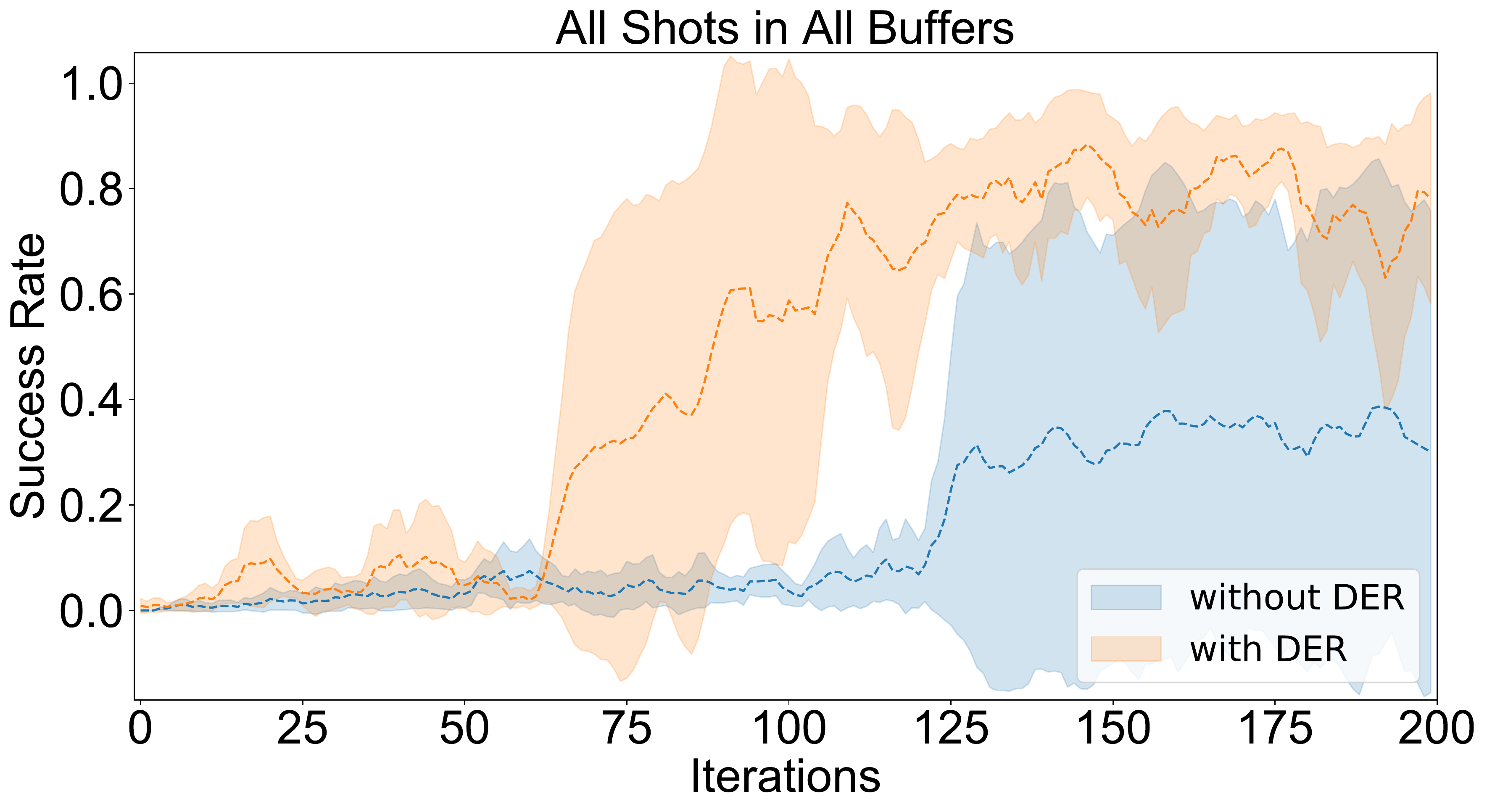} 
    \end{minipage}
    \begin{minipage}{0.38\textwidth}
        \centering
        \includegraphics[width=\textwidth]{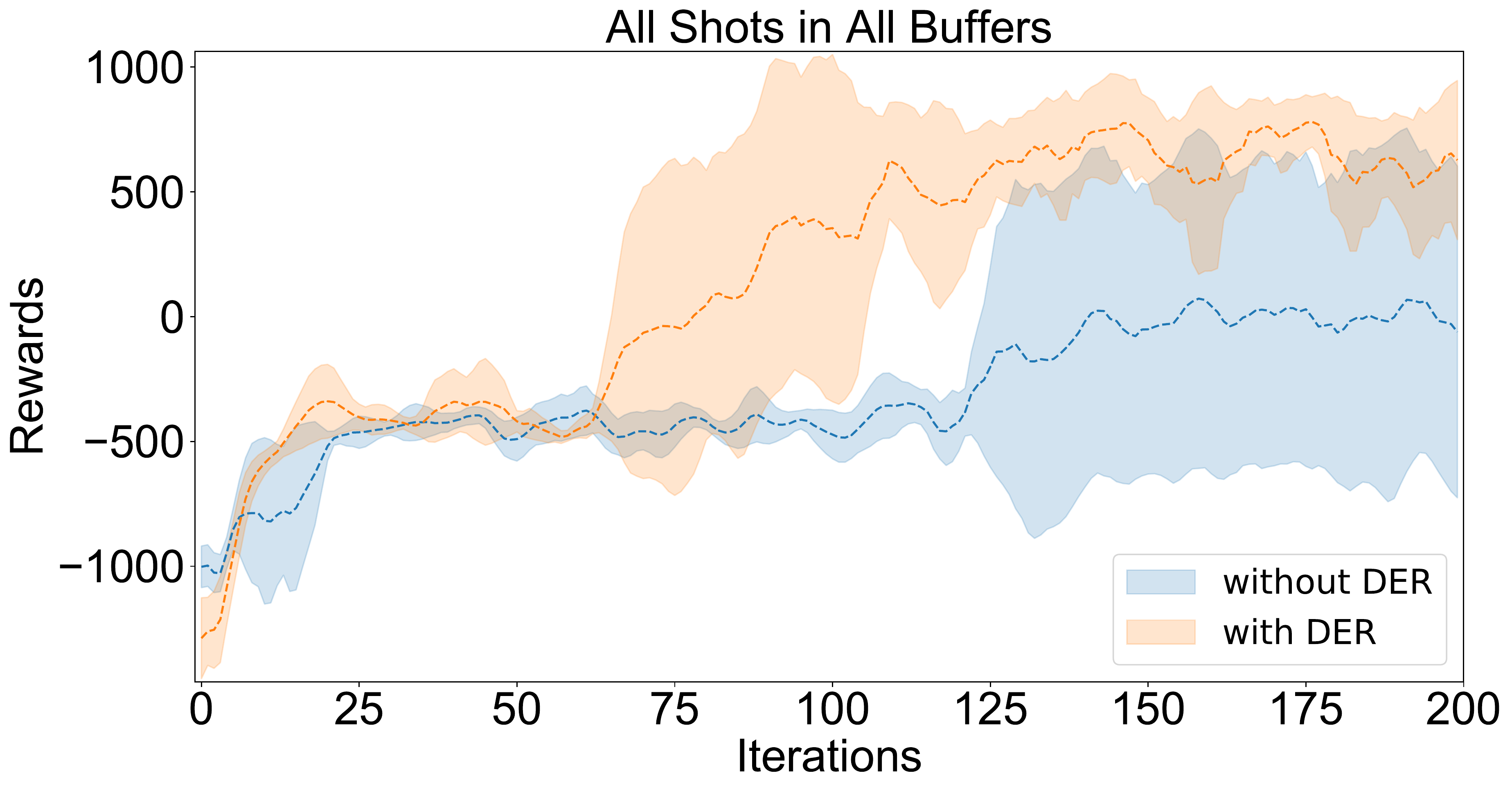} 
    \end{minipage}
    
    \begin{minipage}{0.38\textwidth}
        \centering
        \includegraphics[width=\textwidth]{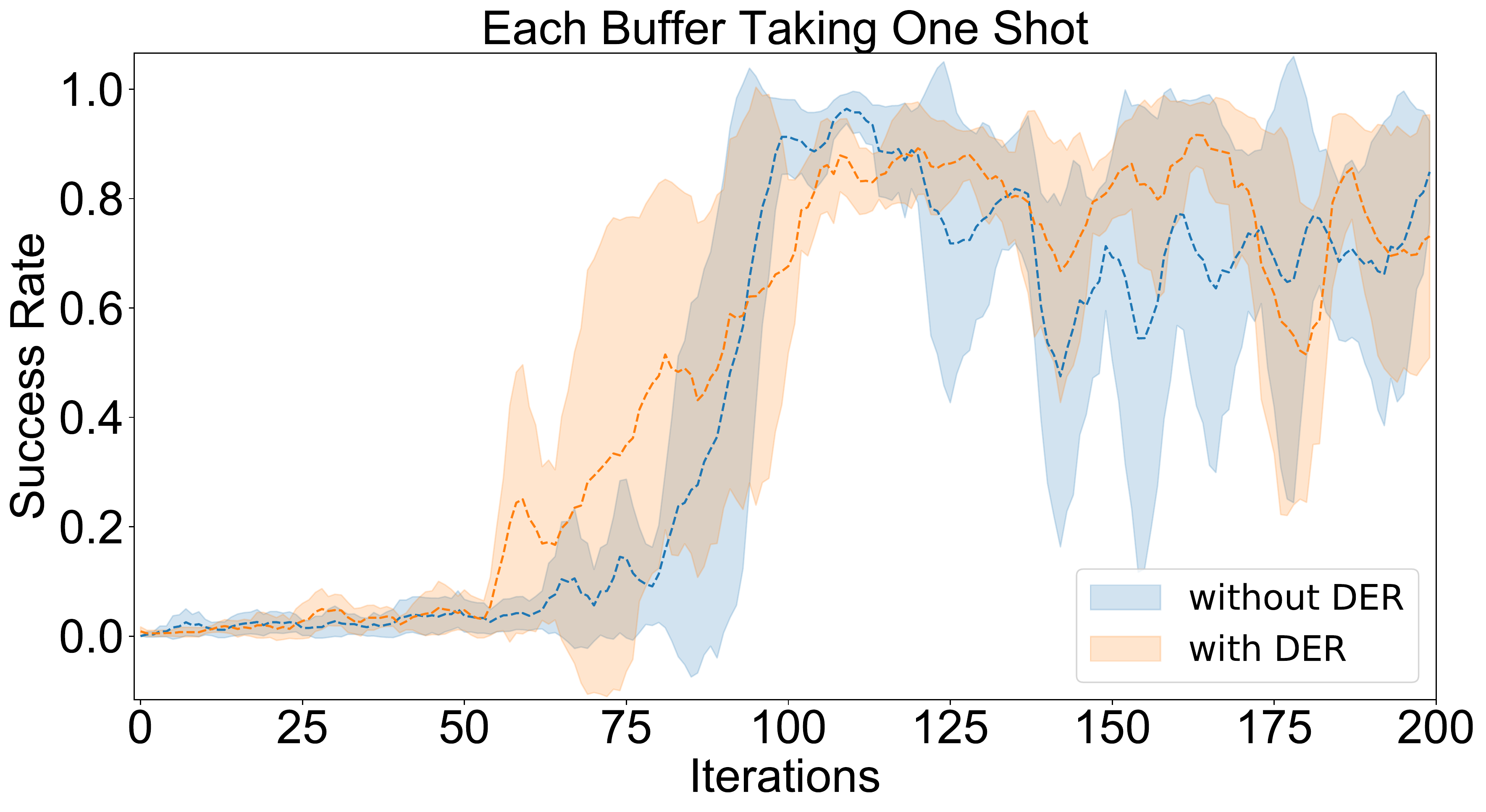} 
    \end{minipage}
    \begin{minipage}{0.38\textwidth}
        \centering
        \includegraphics[width=\textwidth]{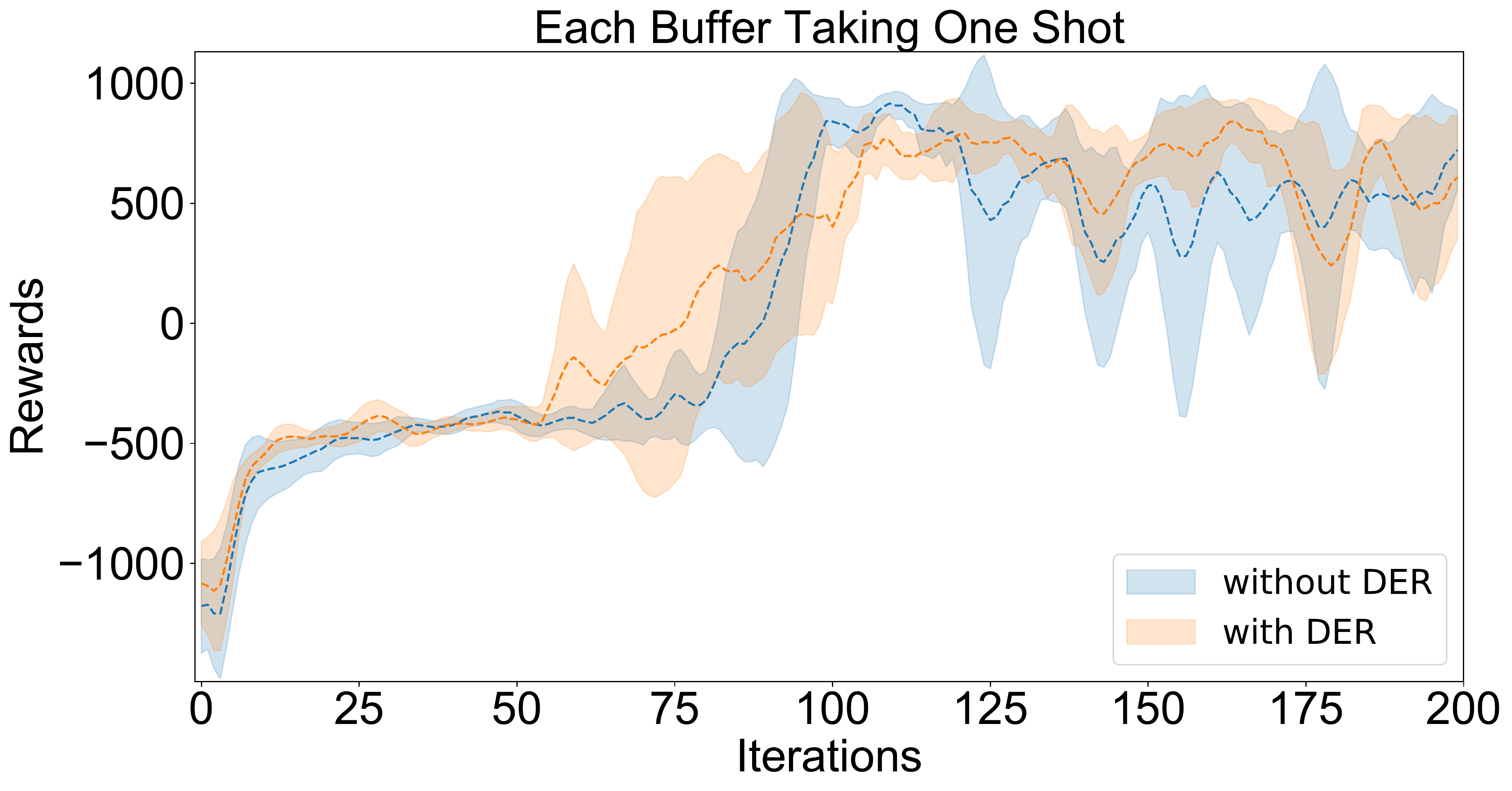}
    \end{minipage}
    
    \caption{Success rate comparison (left) and reward comparison (right) for the peg-in-hole experiments, in which the initial angle of the peg along the z-axis is randomized within [0, $360^\circ$]. Each plot compares the performances of a replay buffer structure with and without DER. Each iteration consists of 50 to 80 episodes and is approximately 200,000 timesteps. The dotted lines show the mean of each iteration across 3 trainings with different random seeds and the shaded areas show the 95\% confidence bound. Each training experiment is terminated at 200 iterations.}
    \label{fig:results_peg}
\end{figure}

\begin{figure}[h]

    \centering
    
    \begin{minipage}{0.38\textwidth}
        \centering
        \includegraphics[width=\textwidth]{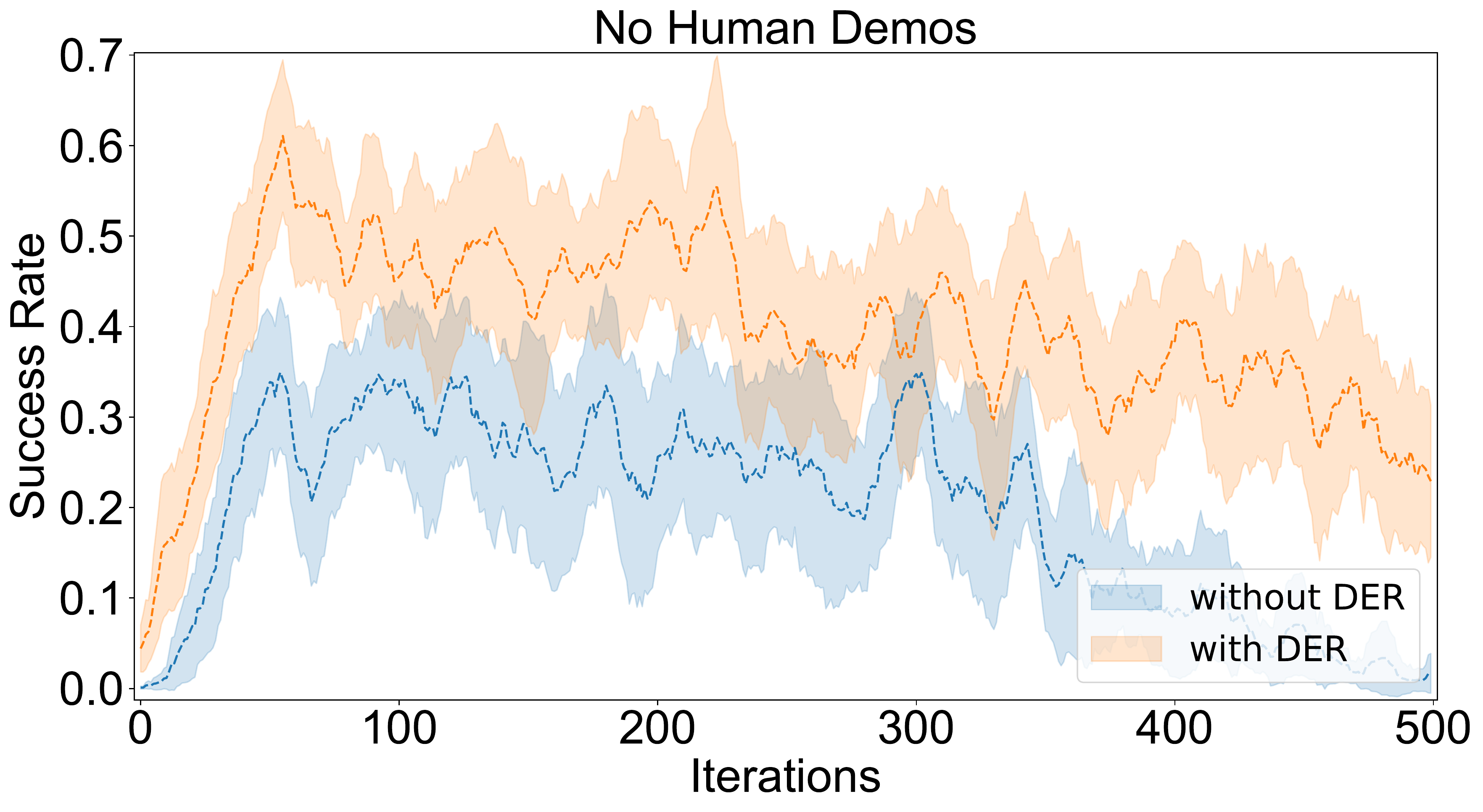} 
    \end{minipage}
    \begin{minipage}{0.38\textwidth}
        \centering
        \includegraphics[width=\textwidth]{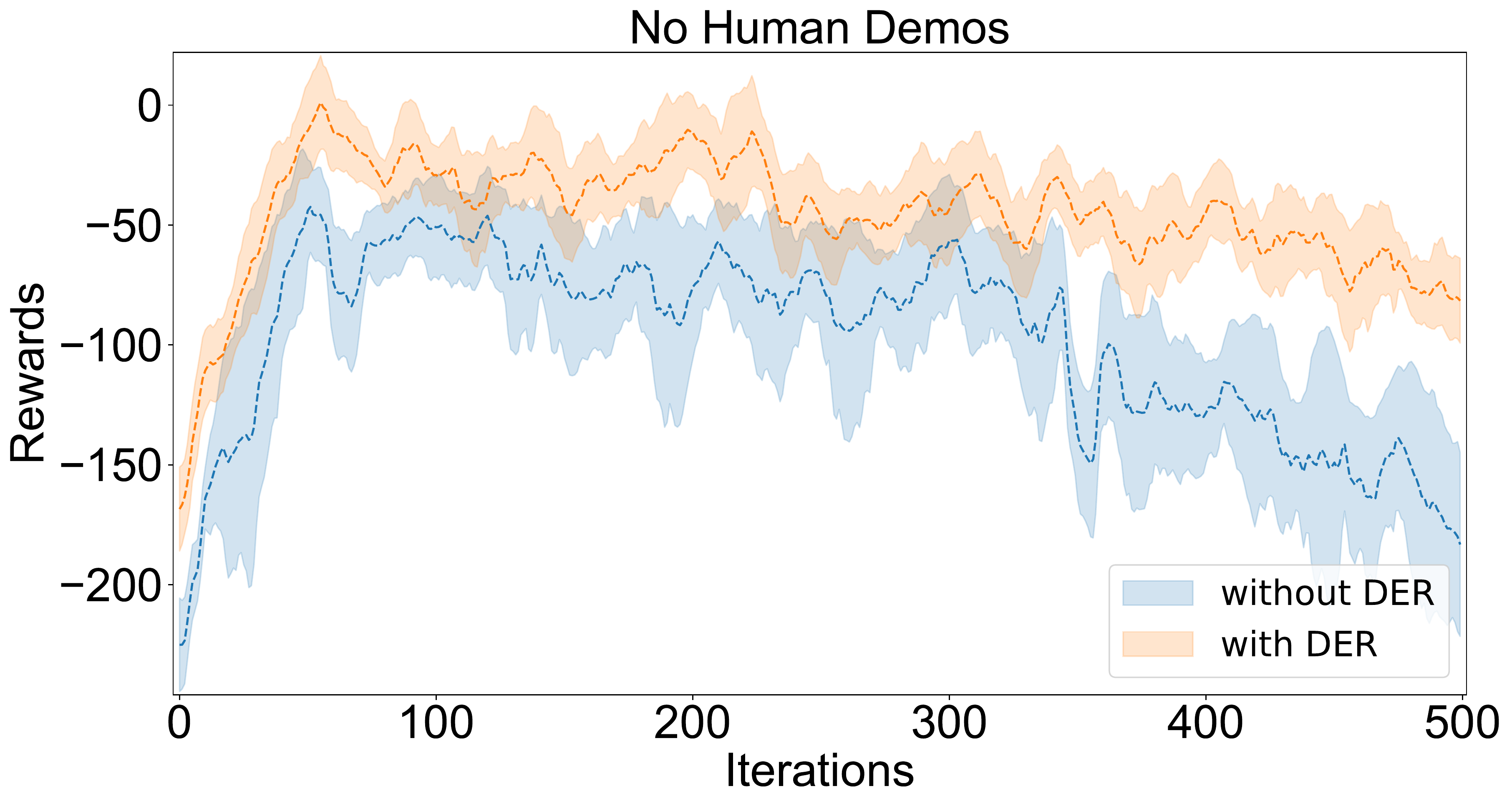} 
    \end{minipage}
    
    \begin{minipage}{0.38\textwidth}
        \centering
        \includegraphics[width=\textwidth]{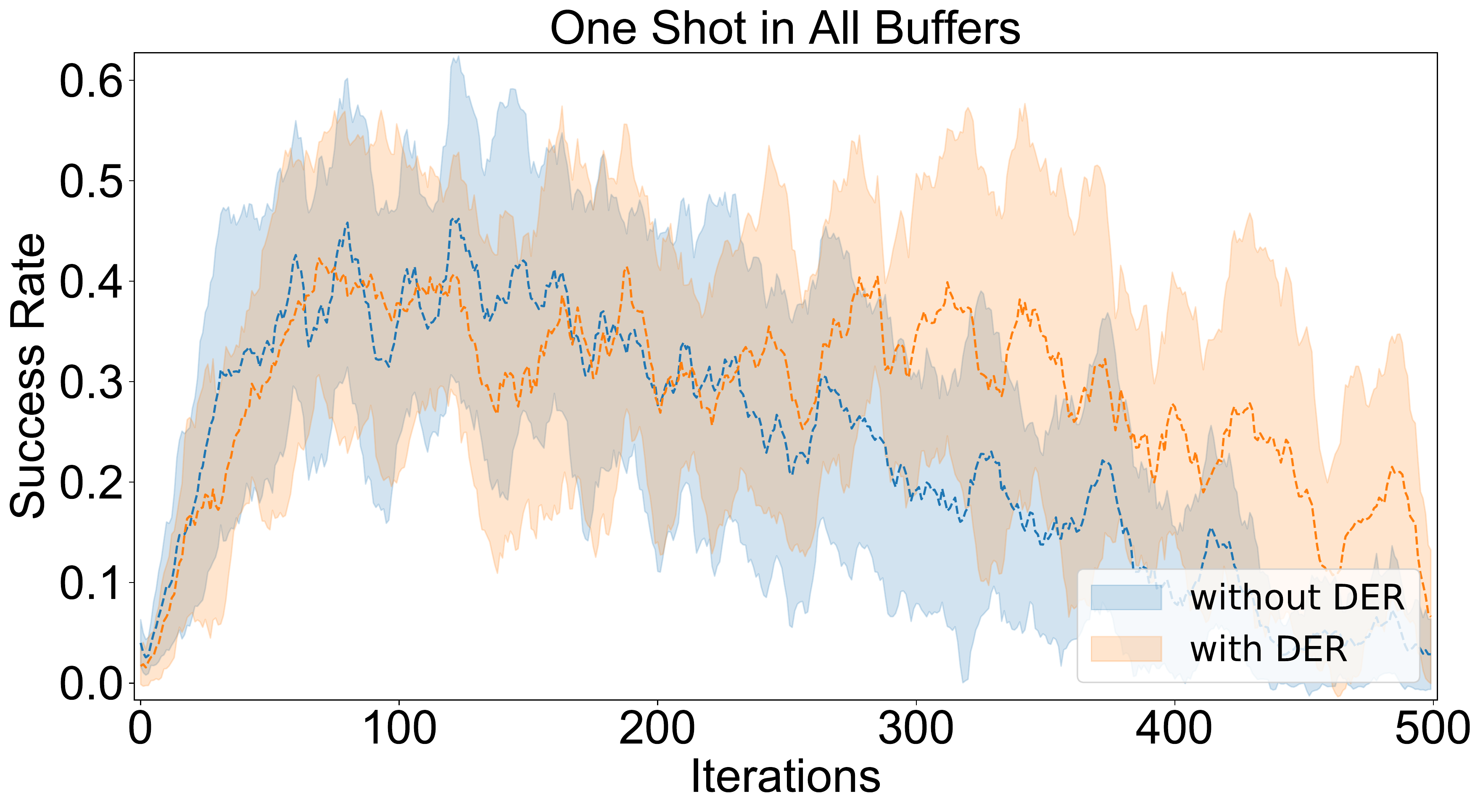} 
    \end{minipage}
    \begin{minipage}{0.38\textwidth}
        \centering
        \includegraphics[width=\textwidth]{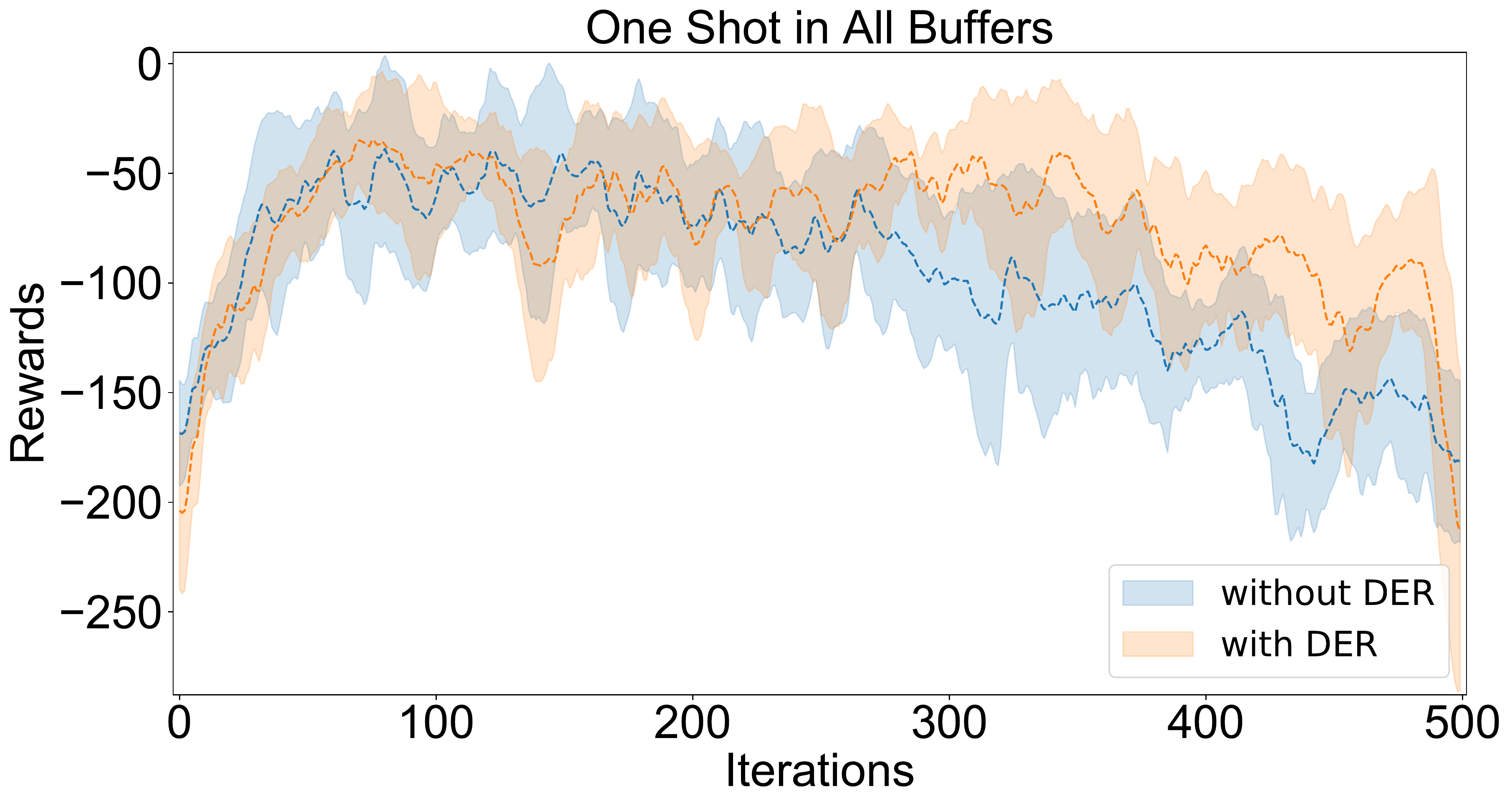}
    \end{minipage}
    
    \begin{minipage}{0.38\textwidth}
        \centering
        \includegraphics[width=\textwidth]{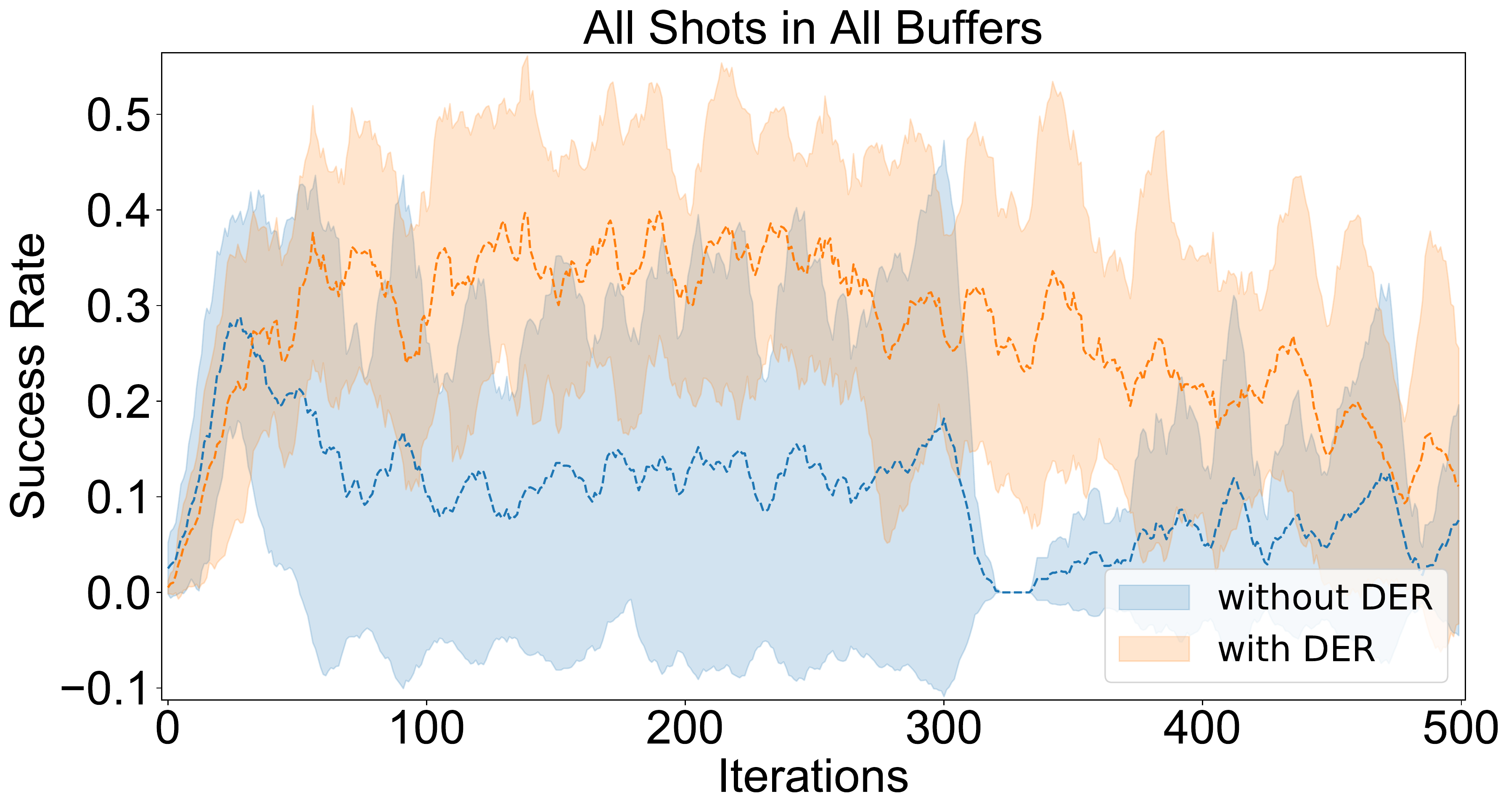} 
    \end{minipage}
    \begin{minipage}{0.38\textwidth}
        \centering
        \includegraphics[width=\textwidth]{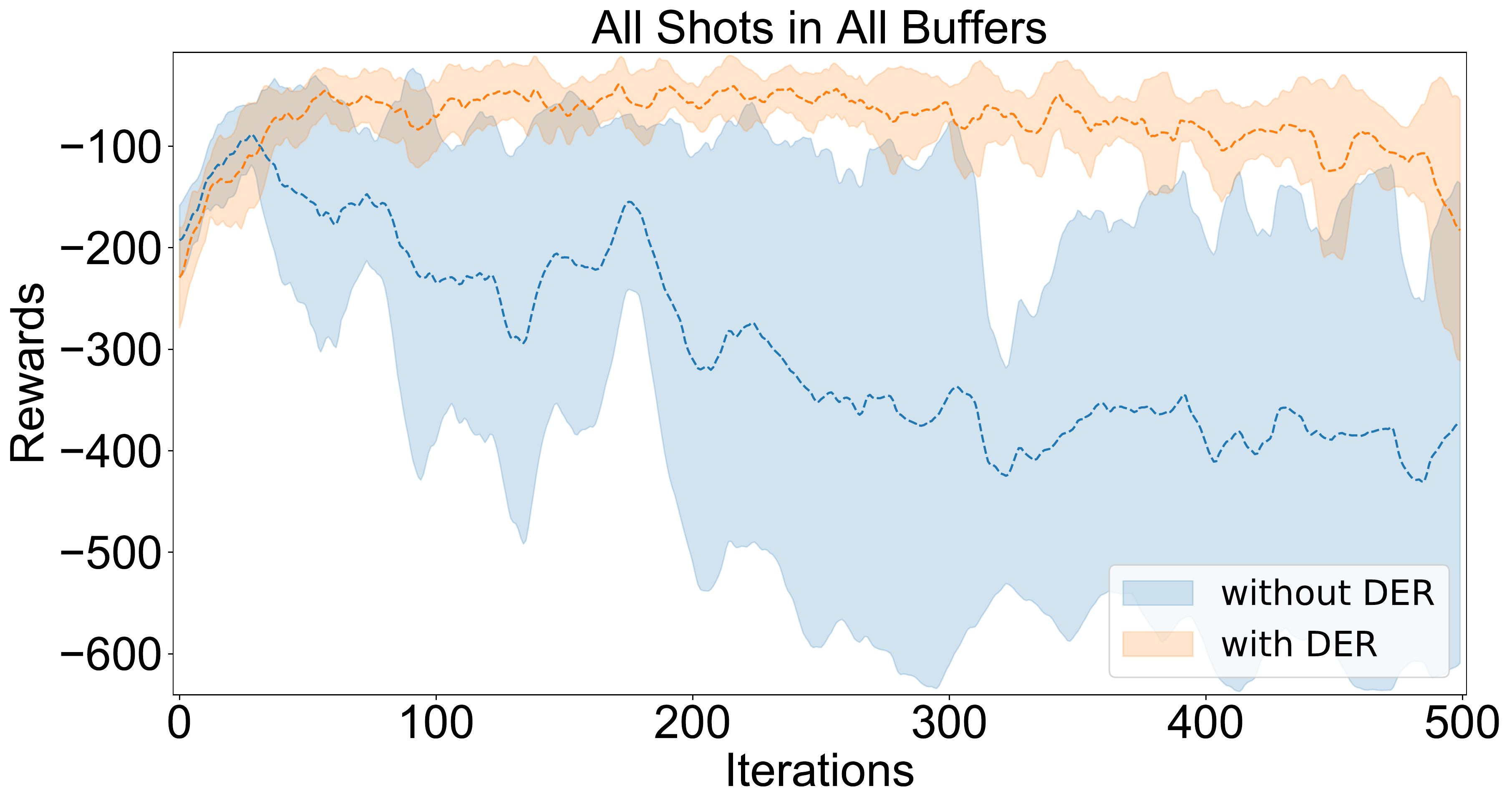} 
    \end{minipage}
    
    \begin{minipage}{0.38\textwidth}
        \centering
        \includegraphics[width=\textwidth]{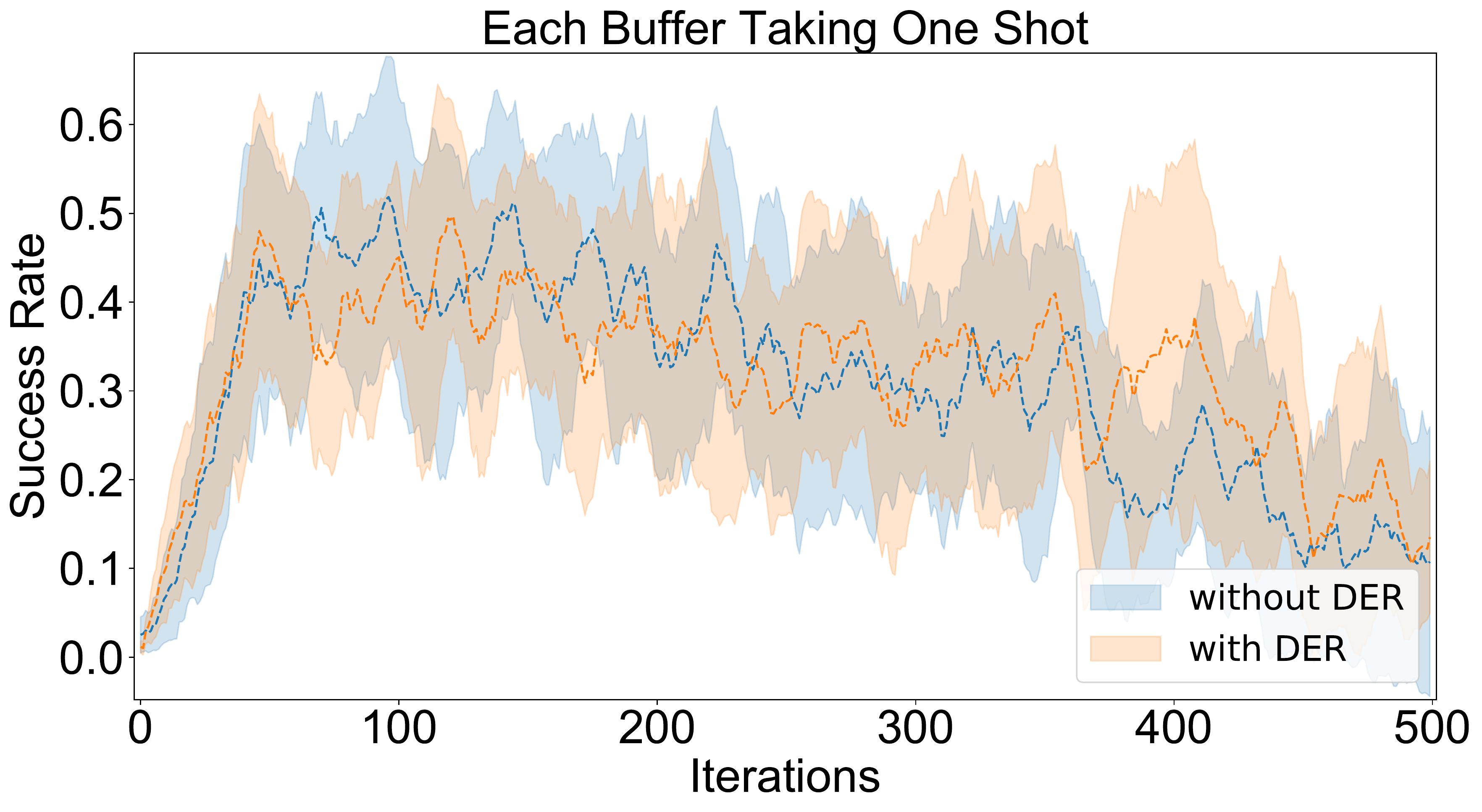}
    \end{minipage}
    \begin{minipage}{0.38\textwidth}
        \centering
        \includegraphics[width=\textwidth]{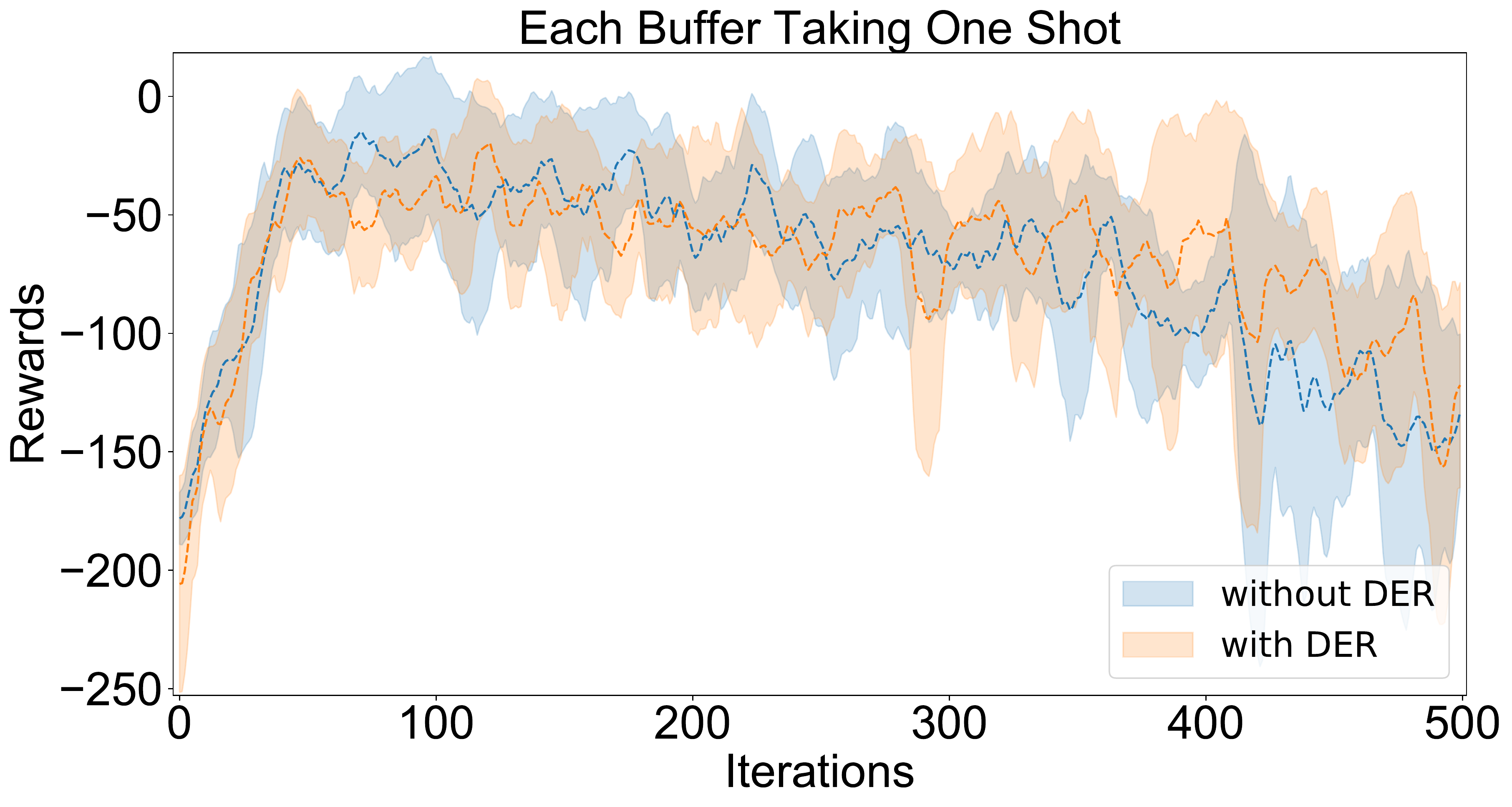}
    \end{minipage}

    \caption{Success rate comparison (left) and reward comparison (right) for the lap-joint experiments, in which the initial angle of the ground timber piece along the z-axis is randomized within [$-2^\circ$, 0] and the initial position [-2mm, 2mm] in both x and y. Each graph compares the performances of a replay buffer structure with and without DER. Each iteration consists of 50 to 80 episodes and is approximately 200,000 timesteps. The dotted lines show the mean of each iteration across 3 trainings with different random seeds and the shaded areas show the 95\% confidence bound. Each training experiment is terminated at 500 iterations.}
    \label{fig:results_lap}
\end{figure}


In order to evaluate how DER affects the performance we evaluate Ape-X DDPG with and without DER on both two tasks. For each task, we conducted eight types of experiments, which are of four different replay-buffer structures with and without DER. Each experiment was performed on an Amazon AWS c5n.9xlarge instance. 

Fig.~\ref{fig:results_peg} shows that DER significantly improves the performance of the peg-in-hole task in most of the buffer structures, including No Human Demos, One Shot in All Buffers, and All Shots in All Buffers. For the latter two buffer structures, the average successful rates of DER are greatly higher than vanilla Ape-X DDPG. For the No-Human-Demos buffer structure, although both algorithms have similar average successful rates by the end of the training, DER is nearly two times as fast at achieving the success rate as vanilla Ape-X DDPG. For the Each-Buffer-Taking-One-Shot buffer structure, with DER and without have similar performance.     

The lap-joint task is more challenging because the timber pieces have straight corners (no chamfer) and tight tolerance (1mm). Hence, as seen in Fig.~\ref{fig:results_lap}, the average success rate of each iteration across different training runs is slightly lower than the peg-in-hole task. Fig.~\ref{fig:results_lap} shows that DER has better performances than vanilla Ape-X DDPG with two of the buffer structures, No Human Demos and All Shots in All Buffers, while the performances of with DER with the other two buffer structures are similar to without DER. It is unclear why DER does not improve the performance with the Each-Buffer-Taking-One-Shot structure in either task. Further studies need to be conducted.   

\subsection{Deployment on a physical robot}
\label{sec: realRobot}
	
After training purely in simulation, we deployed the learned policy of the lap-joint task on a KUKA KR60 industrial robot arm, as shown in Fig.~\ref{fig:tasks}(b). Our hardware setup includes an ATI Delta 6-axis force/torque sensor, two Schunk parallel-jaw grippers, and two pre-fabricated timber pieces with a half notch on each. As discussed in Sec.~\ref{sec: envs}, the observations are force/torque values obtained from the force/torque sensor and pose information of the top timber piece obtained from the robot controller. The actions, linear and angular velocity of the top timber piece, are sent to the robot controller from the policy. The No-Human-Demos buffer structure was used for training the policy, which was successfully deployed on the real robot 3 out of 3 times. We have included the deployment in the video.

\section{Discussion and Future Work}
\label{sec:conclusion}

This paper proposed a novel technique called Dynamic Experience Replay (DER), which improves training efficiency of an off-policy RL algorithm. The technique uses successful episodes generated by RL agents as demonstrations in replay buffers to augment human demonstrations. DER can be seen as a technique of over-sampling the under-represented class from imbalanced data in supervised learning. Our technique can be considered as an add-on feature to an arbitrary off-policy RL algorithm and we experimentally demonstrated that with Ape-X DDPG. 

We showed that DER in both the peg-in-hole and the lap-joint tasks improved training efficiency in comparison to the vanilla Ape-X DDPG algorithm. For occasions where the vanilla RL algorithm failed to solve the task within the given timeframe, DER could either achieve the training goal or largely improve the success rate. We also showed that the learned policy for the lap-joint task can be successfully deployed on the real robot.      

In the future, we would like to evaluate DER on a group of model-free off-policy RL algorithms, such as PPO, and on other assembly tasks. We would also further study DER in terms of hyperparameters, such as the sampling rate and the number of replay buffers. 


 
\clearpage
\acknowledgments{We thank Erin Bradner and Mike Haley for budgetary support of the project; Nicolas Cote for setting up communication and control of the KUKA robot; Matteo Pacher, Aleksandra Anna Apolinarska and Rafael Pastrana for discussions.}


\appendix
\section{Hyperparameter Details}

We used Adam~\citep{kingma2014adam} as the optimizer for both the actor and the critic networks with a learning rate of $10^{-3}$. Instead of using two learning rates, we used two different loss coefficients, 0.1 for the actor and 1.0 for the critic. The target network update frequency is 50,000 and the buffer size is 2,000,000. We allocated 1\% of the buffer size for storing demonstrations, which is 20,000. We used prioritized experience reply and the prioritized replay alpha is 0.5. The sample batch size is 50, the train batch size is 512, and the batch mode is truncate episodes. The minimum per iteration time is 20 seconds. We set soft target updates $\tau$ to 1 as we used target network update frequency for network update. We used mean standard filter as the observation filter. 

For all the peg-in-hole tasks, both the actor and the critic networks have 2 hidden layers with 64 and 64 units. We assigned 5 workers and each of them occupied one logical CPU core. For all the lap-joint tasks, both the actor and the critic networks have 2 hidden layers with 256 and 256 units. We assigned 30 workers and each of them occupied one logical CPU core as well.   

For the rest of the hyperparameters, we inherited directly from RLlib's default setup of Ape-X DDPG.

\bibliography{ms}  

\begin{thebibliography}{28}
\providecommand{\natexlab}[1]{#1}
\providecommand{\url}[1]{\texttt{#1}}
\expandafter\ifx\csname urlstyle\endcsname\relax
  \providecommand{\doi}[1]{doi: #1}\else
  \providecommand{\doi}{doi: \begingroup \urlstyle{rm}\Url}\fi

\bibitem[Lillicrap et~al.(2016)Lillicrap, Hunt, Pritzel, Heess, Erez, Tassa,
  Silver, and Wierstra]{lillicrap2016ddpg}
T.~P. Lillicrap, J.~J. Hunt, A.~Pritzel, N.~Heess, T.~Erez, Y.~Tassa,
  D.~Silver, and D.~Wierstra.
\newblock Continuous control with deep reinforcement learning.
\newblock In \emph{6th International Conference on Learning Representations},
  2016.

\bibitem[Mnih et~al.(2015)Mnih, Kavukcuoglu, Silver, Rusu, Veness, Bellemare,
  Graves, Riedmiller, Fidjeland, Ostrovski, et~al.]{mnih2015dqn}
V.~Mnih, K.~Kavukcuoglu, D.~Silver, A.~A. Rusu, J.~Veness, M.~G. Bellemare,
  A.~Graves, M.~Riedmiller, A.~K. Fidjeland, G.~Ostrovski, et~al.
\newblock Human-level control through deep reinforcement learning.
\newblock \emph{Nature 518, pages 529–533}, 2015.

\bibitem[Horgan et~al.(2018)Horgan, Quan, Budden, Barth-Maron, Hessel,
  Van~Hasselt, and Silver]{horgan2018distributed}
D.~Horgan, J.~Quan, D.~Budden, G.~Barth-Maron, M.~Hessel, H.~Van~Hasselt, and
  D.~Silver.
\newblock Distributed prioritized experience replay.
\newblock \emph{arXiv preprint arXiv:1803.00933}, 2018.

\bibitem[Sutton and Barto(2018)]{sutton2018reinforcement}
R.~S. Sutton and A.~G. Barto.
\newblock \emph{Reinforcement learning: An introduction}.
\newblock MIT press, 2018.

\bibitem[Kober et~al.(2013)Kober, Bagnell, and Peters]{kober2013survey}
J.~Kober, J.~A. Bagnell, and J.~Peters.
\newblock Reinforcement learning in robotics: A survey.
\newblock \emph{The International Journal of Robotics Research 32.11 (2013),
  pp. 1238–1274}, 2013.

\bibitem[Levine and Koltun(2013)]{levine2013guided}
S.~Levine and V.~Koltun.
\newblock Guided policy search.
\newblock In \emph{International Conference on Machine Learning}, pages 1--9,
  2013.

\bibitem[Schulman et~al.(2015)Schulman, Levine, Abbeel, Jordan, and
  Moritz]{schulman2015trust}
J.~Schulman, S.~Levine, P.~Abbeel, M.~Jordan, and P.~Moritz.
\newblock Trust region policy optimization.
\newblock In \emph{International Conference on Machine Learning}, pages
  1889--1897, 2015.

\bibitem[Schulman et~al.(2017)Schulman, Wolski, Dhariwal, Radford, and
  Klimov]{schulman2017proximal}
J.~Schulman, F.~Wolski, P.~Dhariwal, A.~Radford, and O.~Klimov.
\newblock Proximal policy optimization algorithms.
\newblock \emph{arXiv preprint arXiv:1707.06347}, 2017.

\bibitem[Ve{\v{c}}er{\'\i}k et~al.(2017)Ve{\v{c}}er{\'\i}k, Hester, Scholz,
  Wang, Pietquin, Piot, Heess, Roth{\"o}rl, Lampe, and
  Riedmiller]{vevcerik2017leveraging}
M.~Ve{\v{c}}er{\'\i}k, T.~Hester, J.~Scholz, F.~Wang, O.~Pietquin, B.~Piot,
  N.~Heess, T.~Roth{\"o}rl, T.~Lampe, and M.~Riedmiller.
\newblock Leveraging demonstrations for deep reinforcement learning on robotics
  problems with sparse rewards.
\newblock \emph{arXiv preprint arXiv:1707.08817}, 2017.

\bibitem[Lin(1992)]{lin1992self}
L.-J. Lin.
\newblock Self-improving reactive agents based on reinforcement learning,
  planning and teaching.
\newblock \emph{Machine learning}, 8\penalty0 (3-4):\penalty0 293--321, 1992.

\bibitem[Schaul et~al.(2015)Schaul, Quan, Antonoglou, and
  Silver]{schaul2015prioritized}
T.~Schaul, J.~Quan, I.~Antonoglou, and D.~Silver.
\newblock Prioritized experience replay.
\newblock \emph{arXiv preprint arXiv:1511.05952}, 2015.

\bibitem[Andrychowicz et~al.(2017)Andrychowicz, Wolski, Ray, Schneider, Fong,
  Welinder, McGrew, Tobin, Abbeel, and Zaremba]{andrychowicz2017hindsight}
M.~Andrychowicz, F.~Wolski, A.~Ray, J.~Schneider, R.~Fong, P.~Welinder,
  B.~McGrew, J.~Tobin, O.~P. Abbeel, and W.~Zaremba.
\newblock Hindsight experience replay.
\newblock In \emph{Advances in Neural Information Processing Systems}, pages
  5048--5058, 2017.

\bibitem[Chawla et~al.(2002)Chawla, Bowyer, Hall, and
  Kegelmeyer]{chawla2002smote}
N.~V. Chawla, K.~W. Bowyer, L.~O. Hall, and W.~P. Kegelmeyer.
\newblock Smote: synthetic minority over-sampling technique.
\newblock \emph{Journal of artificial intelligence research}, 16:\penalty0
  321--357, 2002.

\bibitem[He et~al.(2008)He, Bai, Garcia, and Li]{he2008adasyn}
H.~He, Y.~Bai, E.~A. Garcia, and S.~Li.
\newblock Adasyn: Adaptive synthetic sampling approach for imbalanced learning.
\newblock In \emph{2008 IEEE International Joint Conference on Neural Networks
  (IEEE World Congress on Computational Intelligence)}, pages 1322--1328. IEEE,
  2008.

\bibitem[Inoue et~al.(2017)Inoue, De~Magistris, Munawar, Yokoya, and
  Tachibana]{inoue2017deep}
T.~Inoue, G.~De~Magistris, A.~Munawar, T.~Yokoya, and R.~Tachibana.
\newblock Deep reinforcement learning for high precision assembly tasks.
\newblock In \emph{2017 IEEE/RSJ International Conference on Intelligent Robots
  and Systems (IROS)}, pages 819--825. IEEE, 2017.

\bibitem[Hochreiter and Schmidhuber(1997)]{hochreiter1997long}
S.~Hochreiter and J.~Schmidhuber.
\newblock Long short-term memory.
\newblock \emph{Neural computation}, 9\penalty0 (8):\penalty0 1735--1780, 1997.

\bibitem[Luo et~al.(2018)Luo, Solowjow, Wen, Ojea, and Agogino]{luo2018deep}
J.~Luo, E.~Solowjow, C.~Wen, J.~A. Ojea, and A.~M. Agogino.
\newblock Deep reinforcement learning for robotic assembly of mixed deformable
  and rigid objects.
\newblock In \emph{2018 IEEE/RSJ International Conference on Intelligent Robots
  and Systems (IROS)}, pages 2062--2069. IEEE, 2018.

\bibitem[Montgomery and Levine(2016)]{montgomery2016guided}
W.~H. Montgomery and S.~Levine.
\newblock Guided policy search via approximate mirror descent.
\newblock In \emph{Advances in Neural Information Processing Systems}, pages
  4008--4016, 2016.

\bibitem[Fan et~al.(2018)Fan, Luo, and Tomizuka]{fan2018guided}
Y.~Fan, J.~Luo, and M.~Tomizuka.
\newblock A learning framework for high precision industrial assembly.
\newblock \emph{arXiv preprint arXiv:1809.08548v3}, 2018.

\bibitem[Luo et~al.(2019)Luo, Solowjow, Wen, Ojea, Agogino, Tamar, and
  Abbeel]{luo2019reinforcement}
J.~Luo, E.~Solowjow, C.~Wen, J.~A. Ojea, A.~M. Agogino, A.~Tamar, and
  P.~Abbeel.
\newblock Reinforcement learning on variable impedance controller for
  high-precision robotic assembly.
\newblock \emph{arXiv preprint arXiv:1903.01066}, 2019.

\bibitem[Todorov and Li(2005)]{todorov2005generalized}
E.~Todorov and W.~Li.
\newblock A generalized iterative lqg method for locally-optimal feedback
  control of constrained nonlinear stochastic systems.
\newblock In \emph{Proceedings of the 2005, American Control Conference,
  2005.}, pages 300--306. IEEE, 2005.

\bibitem[Barth-Maron et~al.(2018)Barth-Maron, Hoffman, Budden, Dabney, Horgan,
  Muldal, Heess, and Lillicrap]{barth2018distributed}
G.~Barth-Maron, M.~W. Hoffman, D.~Budden, W.~Dabney, D.~Horgan, A.~Muldal,
  N.~Heess, and T.~Lillicrap.
\newblock Distributed distributional deterministic policy gradients.
\newblock In \emph{6th International Conference on Learning Representations},
  2018.

\bibitem[Heess et~al.(2017)Heess, Sriram, Lemmon, Merel, Wayne, Tassa, Erez,
  Wang, Eslami, Riedmiller, et~al.]{heess2017emergence}
N.~Heess, S.~Sriram, J.~Lemmon, J.~Merel, G.~Wayne, Y.~Tassa, T.~Erez, Z.~Wang,
  S.~Eslami, M.~Riedmiller, et~al.
\newblock Emergence of locomotion behaviours in rich environments.
\newblock \emph{arXiv preprint arXiv:1707.02286}, 2017.

\bibitem[Adamski et~al.(2018)Adamski, Adamski, Grel, Jedrych, Kaczmarek, and
  Michalewski]{adamski2018distributed}
I.~Adamski, R.~Adamski, T.~Grel, A.~Jedrych, K.~Kaczmarek, and H.~Michalewski.
\newblock Distributed deep reinforcement learning: Learn how to play atari
  games in 21 minutes.
\newblock In \emph{International Conference on High Performance Computing},
  pages 370--388. Springer, 2018.

\bibitem[Liang et~al.(2017)Liang, Liaw, Moritz, Nishihara, Fox, Goldberg,
  Gonzalez, Jordan, and Stoica]{liang2017rllib}
E.~Liang, R.~Liaw, P.~Moritz, R.~Nishihara, R.~Fox, K.~Goldberg, J.~E.
  Gonzalez, M.~I. Jordan, and I.~Stoica.
\newblock Rllib: Abstractions for distributed reinforcement learning.
\newblock \emph{arXiv preprint arXiv:1712.09381}, 2017.

\bibitem[Nair et~al.(2018)Nair, McGrew, Andrychowicz, Zaremba, and
  Abbeel]{nair2018overcoming}
A.~Nair, B.~McGrew, M.~Andrychowicz, W.~Zaremba, and P.~Abbeel.
\newblock Overcoming exploration in reinforcement learning with demonstrations.
\newblock In \emph{2018 IEEE International Conference on Robotics and
  Automation (ICRA)}, pages 6292--6299. IEEE, 2018.

\bibitem[Coumans and Bai(2016)]{coumans2016pybullet}
E.~Coumans and Y.~Bai.
\newblock Pybullet, a python module for physics simulation for games, robotics
  and machine learning.
\newblock \emph{GitHub repository}, 2016.

\bibitem[Kingma and Ba(2014)]{kingma2014adam}
D.~P. Kingma and J.~Ba.
\newblock Adam: A method for stochastic optimization.
\newblock \emph{arXiv preprint arXiv:1412.6980}, 2014.

\end{thebibliography}

\end{document}